\documentclass{article}

\usepackage{microtype}
\usepackage{graphicx}
\usepackage{subfigure}
\usepackage{booktabs} 

\usepackage{hyperref}

\usepackage[accepted]{icml2024}

\usepackage{amsmath}
\usepackage{amssymb}
\usepackage{mathtools}
\usepackage{amsthm}
\usepackage{hyperref}
\usepackage{comment}

\allowdisplaybreaks

\usepackage{pifont}
\newcommand{\cmark}{\ding{51}}%
\newcommand{\xmark}{\ding{55}}%

\usepackage[capitalize,noabbrev]{cleveref}

\theoremstyle{plain}
\newtheorem{theorem}{Theorem}[section]

\newtheorem{lemma}[theorem]{Lemma}
\newtheorem{corollary}[theorem]{Corollary}
\theoremstyle{definition}

\newtheorem{assumption}[theorem]{Assumption}
\theoremstyle{remark}

\usepackage[textsize=tiny]{todonotes}
\usepackage{multirow}
\usepackage{float}

\icmltitlerunning{A Contextual Combinatorial Bandit Approach to Negotiation}

\begin{document}

\twocolumn[
\icmltitle{A Contextual Combinatorial Bandit Approach to Negotiation}




\begin{icmlauthorlist}
\icmlauthor{Yexin Li}{yyy}
\icmlauthor{Zhancun Mu}{comp}
\icmlauthor{Siyuan Qi}{yyy}
\end{icmlauthorlist}

\icmlaffiliation{yyy}{State Key Laboratory of General Artificial Intelligence, BIGAI, Beijing, China}
\icmlaffiliation{comp}{Peking University}

\icmlcorrespondingauthor{Siyuan Qi}{syqi@bigai.ai}

\icmlkeywords{Machine Learning, ICML}

\vskip 0.3in
]



\printAffiliationsAndNotice{} 

\begin{abstract}
Learning effective negotiation strategies poses two key challenges: the exploration-exploitation dilemma and dealing with large action spaces. However, there is an absence of learning-based approaches that effectively address these challenges in negotiation. This paper introduces a comprehensive formulation to tackle various negotiation problems. Our approach leverages contextual combinatorial multi-armed bandits, with the \textit{bandits} resolving the exploration-exploitation dilemma, and the \textit{combinatorial} nature handles large action spaces. Building upon this formulation, we introduce NegUCB, a novel method that also handles common issues such as partial observations and complex reward functions in negotiation. NegUCB is contextual and tailored for full-bandit feedback without constraints on the reward functions. Under mild assumptions, it ensures a sub-linear regret upper bound. Experiments conducted on three negotiation tasks demonstrate the superiority of our approach.
\end{abstract}

\section{Introduction}
\label{introduction}
\begin{figure*}
    \centering
    \includegraphics[width=0.9\textwidth]{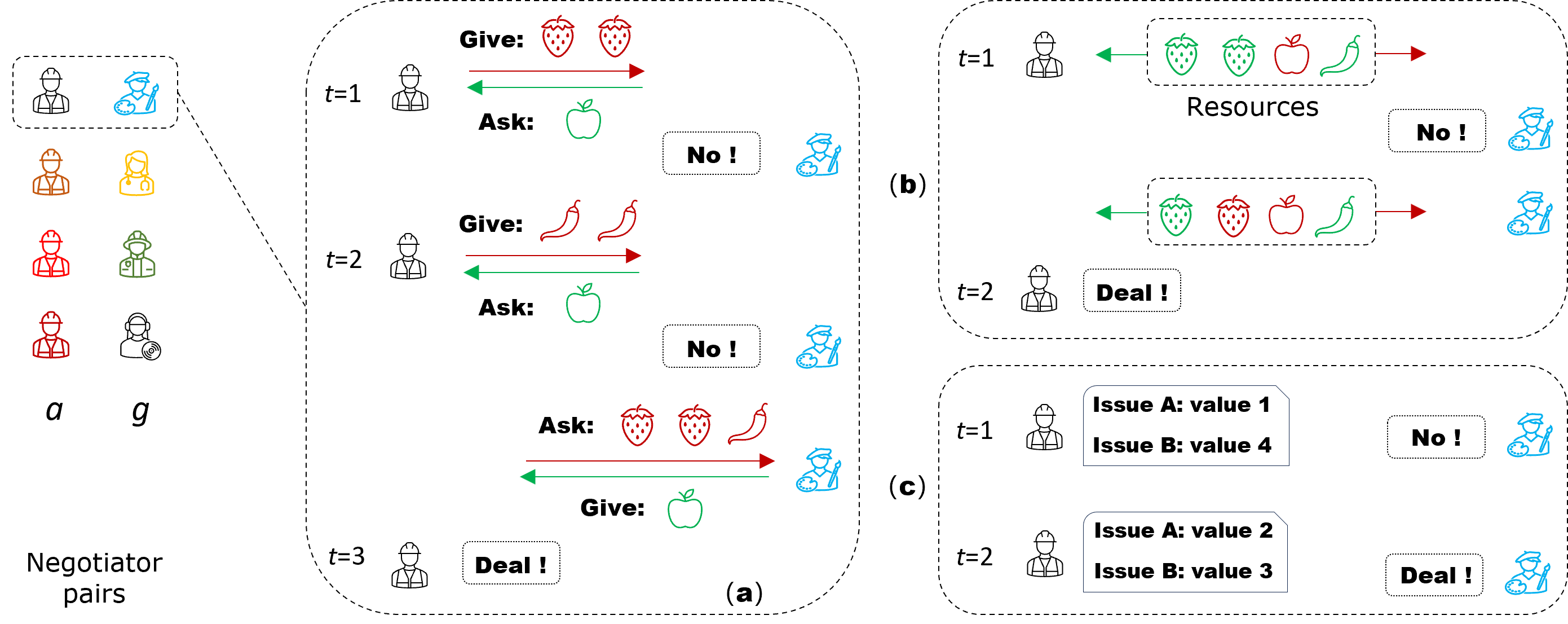}
    \caption{Three typical types of negotiation. Negotiator $a$ is represented with the same icon but in varying colors, indicating the same agent whose state evolves. Negotiator $g$ is depicted with distinct icons and colors, meaning different counterparts. (a) illustrates a trading task, where items in \textbf{Red} signify those that negotiator $a$ gives to $g$, while items in \textbf{Green} indicate those that counterpart $g$ gives to $a$. (b) presents a resource allocation task. Items in green are proposed for allocation to negotiator $a$, while those in red are suggested for assignment to negotiator $g$. Lastly, (c) portrays a multi-issue negotiation task involving two distinct issues, each offering several value choices. Negotiators $a$ and $g$ aim to agree on the values of these two issues.}
    \label{fig:neg_example}

\end{figure*}

Negotiation serves as a fundamental process that underpins interaction among diverse agents across a wide spectrum of domains, ranging from diplomacy~\cite{diplomacy, Cicero} and resource allocation~\cite{deal_or_not_deal, Kris_Cao} to trading~\cite{Anegma}. In these scenarios, an agent, represented as negotiator $a$, engages in negotiation with various counterparts $g$, with its state evolving. At each time step, negotiator $a$ proposes a bid and receives feedback indicating whether the counterpart $g$ accepts or rejects the proposal. Successful acceptance leads to a deal, while rejection leads to termination or further negotiation, possibly with counter-proposals from the counterpart. These negotiations can vary in form, and \cref{fig:neg_example} illustrates three representative negotiation problems: trading, resource allocation, and multi-issue negotiation. As negotiation experiences accumulate, an agent should continuously improve its negotiation ability.

However, effectively exploiting past experiences in subsequent negotiations is challenging in the following aspects. \textbf{Exploration-exploitation dilemma}: As counterparts vary and the agent's state evolves, over-exploiting historical data may result in sub-optimal performance, while excessive exploration may make the counterpart lose patience. Existing works on negotiation~\cite{deal_or_not_deal, Tingwei_Liu, Ayan_Sengupta} tend to neglect exploration, primarily focusing on exploitation, or simply explore by UCT~\cite{MCTS}, without considering observable contexts. \textbf{Large action spaces}: Consider a trading task in which our negotiator possesses items $V_{1}$ while the counterpart holds items $V_{2}$. The potential bid can be any subset of the union $V = V_{1} \cup V_{2}$, resulting in $2^{|V|}$ possible choices. Some studies~\cite{Kris_Cao, RLBOA, Anegma} employ reinforcement learning to acquire negotiation strategies, but they primarily focus on tasks involving action spaces limited to a few hundred discrete actions or low-dimensional continuous action spaces. \textbf{Partial observations}: The profiles of counterparts, including their preferences and desires, cannot be fully observed. Relying solely on observable contexts for negotiation can be ineffective. \textbf{Complicated acceptance functions}: Inferring the likelihood of the counterpart accepting a bid remains challenging, even when their hidden states are known.

In this paper, we formulate negotiation problems using contextual combinatorial multi-armed bandits~\cite{LinUCB, CUCB, C2UCB, ComLinUCB, CC-MAB, DART, ETCG} to address the exploration-exploitation dilemma and handle the large action spaces of combinatorial cardinality. Although negotiation involves a series of actions, unlike in reinforcement learning, where actions may lead to state transitions, bid actions in negotiation do not inherently trigger such transitions. Agents accumulate knowledge about their counterparts through interactions. Consequently, the bandit-based formulation is well-suited for negotiation problems.

In our formulation, an \textbf{arm} denotes an item involved in the negotiation, while a \textbf{super arm} signifies a bid composed of multiple items. The term \textbf{acceptance} is specifically designated to represent the reward, with a value of $1$ assigned when the counterpart accepts the bid and $0$ assigned in the case of rejection. Consequently, our primary objective is the systematic selection of super arms to gain a comprehensive understanding of the expected acceptance of each super arm while ensuring a substantial cumulative benefit in the long run. This formulation involves \textit{full-bandit} feedback, where information regarding the acceptance of individual items within the bid remains inaccessible, and only an aggregate acceptance value for the entire bid is available. Otherwise, the feedback is referred to as \textit{semi-bandit}. Presently, most works \cite{C2UCB, ComLinUCB, CC-MAB, CN-UCB} on combinatorial bandits rely on semi-bandit feedback. Although there are works~\cite{CSAR, DART, ETCG, RGL} that consider full-bandit feedback, they are non-contextual and often subject to specific constraints.

Building upon the above formulation, we propose a contextual algorithm for full-bandit feedback, named \textbf{Neg}otiation \textbf{UCB} (NegUCB), to learn negotiation strategies and adeptly address the exploitation-exploration dilemma and the challenge of large action spaces. Moreover, NegUCB incorporates hidden states to tackle the issue of partial observations and handles diverse acceptance functions through kernel regression \cite{Gaussian_Process, delayed_kernel}. Under mild assumptions, NegUCB's regret upper bound is guaranteed to be sub-linear with respect to the number of negotiation steps and independent of the bid cardinality, distinguishing itself from existing works on either semi-bandit or full-bandit feedback.

In summary, this paper makes three major contributions. First, we provide a comprehensive formulation for diverse types of negotiation problems in \autoref{sec:formulation}. Second, we propose NegUCB to learn negotiation strategies, effectively addressing the prevalent challenges in negotiation in \autoref{sec:negucb}. Lastly, we provide theoretical insights in \autoref{sec:theoretical_analysis} and conduct experiments on representative negotiation tasks in \autoref{sec:experiment}, highlighting the advantages and effectiveness of our method.

\vspace{4pt}
\section{Related Work}
\label{related_work}

\subsection{Negotiation}

Deep reinforcement learning has been applied to learning negotiation strategies. For instance, Rodriguez-Fernandez et al.~\cite{Rodriguez-Fernandez} adopt a DQN-based model~\cite{DQN} to solve the contract negotiation problem characterized by discrete state and action spaces. Lewis et al.~\cite{deal_or_not_deal} combine supervised learning with reinforcement learning to acquire negotiation strategies in a resource allocation task. RLBOA~\cite{RLBOA} discretizes continuous action and state spaces and employs tabular Q-learning to learn bidding strategies, although it may encounter issues related to the curse of dimensionality. ANEGMA~\cite{Anegma} uses actor-critic \cite{actor_critic} to mitigate the dimensionality challenge. Cao et al.~\cite{Kris_Cao} design two communication protocols to explore the emergence of communication when two agents negotiate. However, these approaches struggle to handle large discrete action spaces~\cite{rl_large_action_space} and often give minimal consideration to exploration.

Some studies investigate alternative approaches to negotiation. For instance, Buron et al. learn bidding strategies relying on Monte Carlo tree search~\cite{MCTS}. A decision tree-based negotiation assistant ~\cite{Tingwei_Liu} is specifically designed to predict prices in a car trading platform. Sengupta et al.~\cite{Ayan_Sengupta} demonstrate a transfer learning-based solution to adapt base negotiation strategies to new counterparts rapidly. Cicero~\cite{Cicero} achieves mastery in the game of \textit{Diplomacy} by integrating reinforcement learning with a language model. Nevertheless, these approaches deal with highly specific problems or issues in negotiation, yet they have not effectively tackled the prevalent challenges discussed above.

\vspace{5pt}
\subsection{Multi-Armed Bandits}

LinUCB~\cite{LinUCB} has been introduced to formulate \textit{recommendation} as a contextual bandit problem, assuming linearity in the reward concerning user and item contexts. It has demonstrated effective performance in \textit{recommendation} and guarantees a sub-linear regret bound~\cite{LinUCB_proof}. FactorUCB~\cite{FactorUCB} also makes a linearity assumption but considers hidden features alongside the observable contexts, leading to an improved click rate in \textit{recommendation}. To overcome the linearity assumption in contextual bandits, KernelUCB~\cite{KernelUCB, GP_UCB} transforms contexts into a high-dimensional space and applies LinUCB in this new space. Neural-UCB~\cite{Neural-UCB} attempts to leverage deep neural networks to capture the relationship between contexts and rewards. However, its computational complexity makes it challenging to generalize to real tasks. 

\vspace{3pt}
CUCB~\cite{CUCB} establishes a general framework for combinatorial multi-armed bandits. C2UCB~\cite{C2UCB} and ComLinUCB~\cite{ComLinUCB} incorporate contexts into combinatorial bandits based on the same linearity assumption as LinUCB. CC-MAB~\cite{CC-MAB} focuses on problems with volatile arms and submodular reward functions. CN-UCB~\cite{CN-UCB} employs neural networks to address contextual combinatorial bandit problems, facing the computational limitation as Neural-UCB. However, these algorithms operate within semi-bandit feedback. Another relevant setting is the full-bandit feedback, in which rewards for individual arms are inaccessible. Algorithms designed for full-bandit feedback include CSAR~\cite{CSAR}, DART~\cite{DART}, ETCG~\cite{ETCG}, and RGL~\cite{RGL}. However, their reward functions adhere to linearity or sub-modularity, and none of them consider contexts. In contrast, NegUCB is contextual, combinatorial, and tailored for full-bandit feedback without constraints on the reward functions. A comparative analysis is presented in \cref{tab:algorithms_comparison}.

\begin{table}[t]
\caption{Comparison between NegUCB and representative multi-armed bandit algorithms. \textbf{Contextual}: consider contexts. \textbf{Combinatorial}: consider super arms consisting of multiple basis arms. \textbf{Partial}: contexts are partially observable. \textbf{Non-linear}: non-linear reward functions w.r.t. contexts. \textbf{Full-bandit}: rewards of basis arms are not available. \textit{Blank} means the attribute does not apply.}
\label{tab:algorithms_comparison}
\begin{center}
    \scalebox{0.67}{
    \begin{tabular}{lccccr}
        \toprule
        Algorithm  & Contextual & Combinatorial & Partial & Non-Linear & Full-bandit \\
        \midrule
        LinUCB & \cmark & \xmark & \xmark & \xmark & \\
        FactorUCB & \cmark & \xmark & \cmark & \xmark & \\
        KernelUCB  & \cmark & \xmark & \xmark & \cmark & \\
        Neural-UCB  & \cmark & \xmark & \xmark & \cmark & \\
        \midrule
        
        CUCB  & \xmark & \cmark &  & & \xmark \\
        C2UCB  & \cmark & \cmark & \xmark & \xmark & \xmark \\
        ComLinUCB  & \cmark & \cmark & \xmark & \xmark & \xmark \\
        CC-MAB  & \cmark & \cmark & \xmark & \cmark & \xmark \\
        CN-UCB  & \cmark & \cmark & \xmark & \cmark & \xmark \\
        CSAR & \xmark & \cmark & & & \cmark \\
        DART & \xmark & \cmark & & & \cmark \\
        ETCG & \xmark & \cmark & & & \cmark \\
        RGL & \xmark & \cmark & & & \cmark \\

        \midrule
        \textbf{NegUCB} & \cmark & \cmark & \cmark & \cmark & \cmark \\
        
        \bottomrule
    \end{tabular}
    }

\vspace{-5pt}
\end{center}
\end{table}

\section{Methodology}
\label{methodology}

\begin{figure*}[t!]
    \centering
    \includegraphics[width=0.85\textwidth]{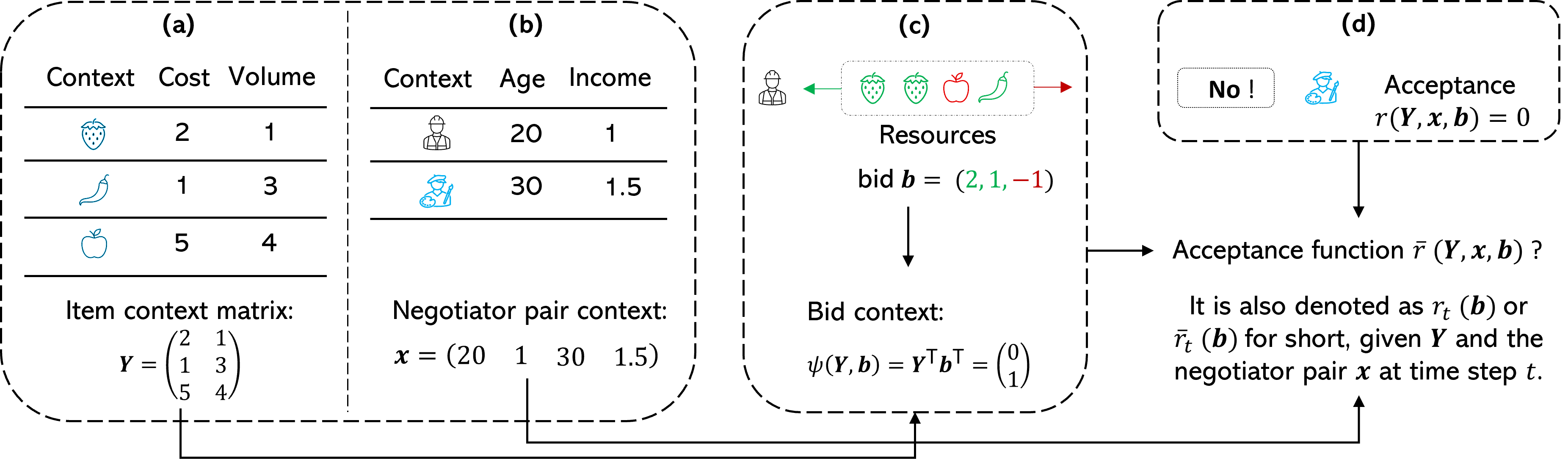}
    \caption{Acceptance function of a resource allocation task. (a) and (b) describe the contexts of the items and the current negotiator pair, respectively. (c) provides an example illustrating how the bid can be defined and how to extract the bid context. (d) depicts the acceptance label. The goal is to approximate the acceptance function $\bar{r}: (\boldsymbol{Y}, \boldsymbol{x}, \boldsymbol{b}) \mapsto r$ using historical negotiation data.}
    \label{fig:NegUCB}
    \vspace{-10pt}
\end{figure*}

Unless otherwise specified, uppercase symbols represent sets, bold uppercase symbols denote matrices, bold lowercase symbols represent vectors, and lowercase symbols denote scalars or functions. $\boldsymbol{I}_{d}$ refers to an identity matrix with dimensions $d \times d$, and $\boldsymbol{0}_{d}$ represents a zero vector of size $d \times 1$. Kronecker product is denoted as $\otimes$. Frobenius norm of a matrix and the $l_{2}$ norm of a vector are respectively denoted as $\left \| \boldsymbol{X} \right \|$ and $\left \| \boldsymbol{x} \right \|$. Mahalanobis norm of a column vector $\boldsymbol{x}$ based on matrix $\boldsymbol{A}$ is denoted as $\left \| \boldsymbol{x} \right \|_{\boldsymbol{A}} = \sqrt{\boldsymbol{x}^{\mathsf{T}} \boldsymbol{A} \boldsymbol{x}}$. $\textup{vec}(\boldsymbol{A})$ is the vectorization operator of matrix $\boldsymbol{A}$.

\subsection{Negotiation Formulation}
\label{sec:formulation}
In this section, we provide a comprehensive formulation that applies to various types of negotiation problems. First, we outline the negotiation framework, detailing the strategy for making proposals when it is our turn to bid and the criteria for deciding whether to accept or reject a bid from the counterpart. Next, we formulate the critical component within this framework.

\subsubsection{Negotiation Framework}
Denote the pool of the counterpart negotiators as $U$, with a cardinality of $|U| = m$. The item pool is represented as $V$, where $|V| = n$. It is essential to acknowledge that negotiation with a new counterpart may occur at any time, leading to an increase in $m$ over time. Additionally, new items may be added to $V$. Without loss of generality, we assume these two pools $U$ and $V$ to be constant. At time step $\tau$, our negotiator has a valid bid set $B_{\tau}$ encompassing all feasible bids it can propose at this time. For example, in a trading task, a bid specifies which items our negotiator proposes to give to the counterpart and which items it requests in return. The validity of a bid is determined by whether our negotiator possesses the items it proposes to give. More designation of bids under various negotiation scenarios will be elaborated later.

For a valid bid $\boldsymbol{b} \in B_{\tau}$ of our negotiator at time $\tau$, the acceptance function $r_{\tau}$ assesses whether the counterpart may accept or reject the bid, denoted by $r_{\tau} (\boldsymbol{b}) = 1$ or $r_{\tau} (\boldsymbol{b}) = 0$. This function is unknown and needs to be learned. Additionally, there exists a benefit function $f_{\tau}$ that measures the potential benefit of the bid to our negotiator, a metric highly dependent on the specific problem. Consequently, the optimal bid for our negotiator to propose at time step $\tau$ is determined by \autoref{eq:bid eq.}, aiming to maximize its expected benefit. During negotiation, if our negotiator intends to propose a bid, it chooses a valid bid using \autoref{eq:bid eq.}. Supposing our negotiator receives a bid $\boldsymbol{b}$ from the counterpart, it is evident that $r_{\tau} (\boldsymbol{b}) = 1$; thus, our negotiator decides whether to accept the bid by evaluating if the bid is valid and optimal after setting $r_{\tau} (\boldsymbol{b}) = 1$.

\begin{small}
    \begin{equation}
    \boldsymbol{b}_{\tau}^{*} = \mathop{\arg\max} r_{\tau} (\boldsymbol{b}) \times f_{\tau} (\boldsymbol{b})
    \label{eq:bid eq.}
\end{equation}
\end{small}

\subsubsection{Acceptance Function}

As the benefit function $f_{\tau}$ is dependent on the specific problem and crafted manually, it is not the primary focus of this work. Instead, we focus on learning the acceptance function $r_{\tau}$ using contextual combinatorial multi-armed bandits~\cite{CUCB, C2UCB, ComLinUCB, CC-MAB, ETCG}, where \textbf{arms} represent items in $V$, \textbf{super arms} denote bids, and rewards are the \textbf{acceptance} labels. Consequently, the objective is to iteratively put forth beneficial bids to understand the expected acceptance of each bid by various counterparts while ensuring a substantial cumulative benefit in the long run. In the following, we review the details based on \cref{fig:NegUCB}.

Items in pool $V$ have contexts denoted as row vectors $\left \{ \boldsymbol{y}_{w}|w=1, 2, ..., n \right \}$, collectively forming an item context matrix $\boldsymbol{Y} = [\boldsymbol{y}_{1}; \boldsymbol{y}_{2}; ...; \boldsymbol{y}_{n}]$, as depicted in \cref{fig:NegUCB} (a). At time step $\tau$, our negotiator and the counterpart form a negotiator pair, characterized by contexts denoted as a row vector $\boldsymbol{x}_{\tau}$, as depicted in \cref{fig:NegUCB} (b). It is worth noting that $\boldsymbol{x}_{\tau}$ corresponds to one of the $m$ counterparts in $U$. In other words, it is a row of the negotiator context matrix $\boldsymbol{X} = [\boldsymbol{x}_{1}; \boldsymbol{x}_{2}; ...; \boldsymbol{x}_{m}]$. In this work, we use $\boldsymbol{x}_{\tau}$ or $\boldsymbol{x}_{w}, w = 1, 2, ..., m$, interchangeably to either highlight the time step or the counterpart index. Addressing the partial observation issue, we assume hidden states $\boldsymbol{U} = [\boldsymbol{u}_{1}; \boldsymbol{u}_{2}; ...; \boldsymbol{u}_{m}]$ for the $m$ negotiator pairs. Then the acceptance function $r_{\tau}$ at time step $\tau$ is estimated through \autoref{eq:accept eq.}, where $\boldsymbol{\Theta}$ in the first term represents the function parameters, $\boldsymbol{u}_{\tau}$ in the second term signifies the hidden state of the current negotiator pair. Specifically, the first term estimates the partial acceptance of the bid based on observed contexts, while the second term evaluates the partial acceptance of the bid based on hidden states.

\begin{small}
    \begin{align}
            \label{eq:accept eq.}
            & \bar{r}_{\tau} (\boldsymbol{b}_{\tau}) = \phi(\boldsymbol{x}_{\tau}) \boldsymbol{\Theta} \left \langle \boldsymbol{Y}, \boldsymbol{b}_{\tau} \right \rangle + \boldsymbol{u}_{\tau} \left \langle \boldsymbol{Y}, \boldsymbol{b}_{\tau} \right \rangle \\[10pt]
            \label{eq:item eq.}
            & \left \langle \boldsymbol{Y}, \boldsymbol{b}_{\tau} \right \rangle = \phi \circ \psi (\boldsymbol{Y}, \boldsymbol{b}_{\tau})
\end{align}
\end{small}

In the first term of \autoref{eq:accept eq.}, function $\phi$ transforms context $\boldsymbol{x}_{\tau}$ into a $h$-dimensional space $\mathcal{H}$ where $h$ can be infinite. $\left \langle \boldsymbol{Y}, \boldsymbol{b}_{\tau} \right \rangle$ is expressed in \autoref{eq:item eq.}, where function $\psi$ extracts the context of bid $\boldsymbol{b}_{\tau}$ from the item context matrix $\boldsymbol{Y}$. A possible example of $\psi$ is provided in \cref{fig:NegUCB} (c). Following this, function $\phi$ further transforms the bid context into a high-dimensional representation within space $\mathcal{H}$. Specifically, function $\phi$ transforms contexts into high-dimensional representations, allowing the acceptance function to operate non-linearly concerning the observed contexts.

Given historical negotiation data from step $1, 2, ..., \tau$, we aim to optimize the following objective function to derive functions $\psi$ and $\phi$, parameters $\boldsymbol{\Theta}$ and hidden states $\boldsymbol{U}$, then use them for the subsequent time step $\tau + 1$. $\lambda_{1}$ and $\lambda_{2}$ are hyper-parameters for the regularization terms, $r_{t}$ represents the actual acceptance at time $t$, as depicted in \cref{fig:NegUCB} (d), and $\bar{r}_{t}$ denotes the acceptance estimated by \autoref{eq:accept eq.}.

\begin{small}
    \begin{equation}
     \min_{\psi, \phi, \boldsymbol{\Theta}, \boldsymbol{U}} \mathcal{L} = \sum_{t=1}^{\tau} \left | \bar{r}_{t} - r_{t} \right |^{2} + \lambda_{1} \left \| \boldsymbol{\Theta} \right \|^{2} + \lambda_{2} \left \| \boldsymbol{U} \right \|^{2}
    \label{eq:obj}
    \end{equation}
\end{small}

\subsection{Negotiation UCB}
\label{sec:negucb}
In this subsection, building upon the above formulation, we introduce the NegUCB algorithm, a simple yet effective approach. In this algorithm, bids are represented as indicator vectors indicating the items involved in each bid. Please refer to \autoref{app:bid_design} for detailed examples.

Since simultaneously deriving the functions $\phi$ and $\psi$, as well as the parameters $\boldsymbol{\Theta}$ and $\boldsymbol{U}$ is challenging, we assume the format of the function $\psi$ in \cref{ass:psi}. In \autoref{subsubsec:parameters}, we provide the closed-form solutions for $\boldsymbol{\Theta}$ and $\boldsymbol{U}$, which depend on function $\phi$. In \autoref{subsubsec:phi_function}, we use kernel regression \cite{Gaussian_Process} to eliminate the dependency on function $\phi$. At last, we summarize the NegUCB algorithm in \autoref{subsubsec:NegUCB_alg}. 

\begin{assumption}
    If the contexts of items are characterized by their basic features, the extraction function $\psi$ in \autoref{eq:psi} can accurately capture the context of bid $\boldsymbol{b}_{\tau}$. In other words, it encompasses substantial information about the items included in the bid.

    \vspace{-5pt}
    \begin{small}
        \begin{equation}
        \psi(\boldsymbol{Y}, \boldsymbol{b}_{\tau}) = \boldsymbol{Y}^{\mathsf{T}} \boldsymbol{b}_{\tau}^{\mathsf{T}}
        \label{eq:psi}
    \end{equation}
    \end{small}
    
\label{ass:psi}
\end{assumption}

\vspace{-2pt}
Despite the linearity assumption on $\psi$, the acceptance function is non-linear because of the transforming function $\phi$.

\subsubsection{Parameters}
\label{subsubsec:parameters}
Obviously, the objective function $\mathcal{L}$ is not jointly convex concerning both $\boldsymbol{\Theta}$ and $\boldsymbol{U}$. However, it is convex concerning one parameter if the other one is fixed. Therefore, we employ an alternative least square optimization approach, iterating the calculation of one parameter with a closed-form solution while keeping the other parameter fixed. Based on \cref{ass:psi}, the closed-form solution for $\boldsymbol{\Theta}$ is as \autoref{eq:theta}, while that for $\boldsymbol{U}$ is as \autoref{eq:u}.

\vspace{-5pt}
\begin{small}
    \begin{align}
    \label{eq:theta}
    & \text{vec}(\boldsymbol{\Theta }) = (\boldsymbol{A}_{\tau}^{\mathsf{T}} \boldsymbol{A}_{\tau} + \lambda_{1} \boldsymbol{I}_{h^{2}})^{-1} \boldsymbol{A}_{\tau}^{\mathsf{T}} (\boldsymbol{r}_{\tau} - \boldsymbol{D}_{\tau} \text{vec}(\boldsymbol{U})) \\[5pt]
    \label{eq:u}
    & \text{vec}(\boldsymbol{U}) = (\boldsymbol{D}_{\tau}^{\mathsf{T}} \boldsymbol{D}_{\tau} + \lambda_{2} \boldsymbol{I}_{mh})^{-1} \boldsymbol{D}_{\tau}^{\mathsf{T}} (\boldsymbol{r}_{\tau} - \boldsymbol{A}_{\tau} \text{vec}(\boldsymbol{\Theta})) 
    \end{align}
\end{small}

Rows of matrices $\boldsymbol{A}_{\tau}$ and $\boldsymbol{D}_{\tau}$ are samples as $\phi(\boldsymbol{b}_{t} \boldsymbol{Y}) \otimes \phi(\boldsymbol{x}_{t})$ and $\phi(\boldsymbol{b}_{t} \boldsymbol{Y}) \otimes \boldsymbol{p}_{t}$ where $\boldsymbol{p}_{t} \in R^{1 \times m}$ is a one-hot vector representing the counterpart index at time step $t = 1, 2, ..., \tau$.  It is evident that the solutions for parameters $\boldsymbol{\Theta}$ and $\boldsymbol{U}$ are contingent on the transformation function $\phi$, which can take various forms, such as polynomial functions, neural networks, etc., and thus needs to be learned.

\subsubsection{Transformation function}
\label{subsubsec:phi_function}
Given the limited amount of negotiation data with various counterparts, learning $\phi$ becomes intractable if it involves many parameters, such as in the case of neural networks. In NegUCB, we utilize Reproducing Kernel Hilbert Spaces within kernel functions to avoid the need for learning $\phi$, enhancing efficiency. Moreover, since iterating among three components, i.e., learning $\phi$, $\boldsymbol{U}$, and $\boldsymbol{\Theta}$, is highly unstable, NegUCB iterates between learning $\boldsymbol{U}$ and $\boldsymbol{\Theta}$, significantly improving the learning stability.

Corresponding to matrices $\boldsymbol{A}_{\tau}$ and $\boldsymbol{D}_{\tau}$ dependent on function $\phi$, we define matrices $\boldsymbol{K}_{\tau}$ and $\boldsymbol{Z}_{\tau}$. Each entry $(\boldsymbol{K}_{\tau})_{t, j}$ and $(\boldsymbol{Z}_{\tau})_{t, j}$ are the dot product of the $t$-th and $j$-th samples of $\boldsymbol{A}_{\tau}$ and $\boldsymbol{D}_{\tau}$, respectively. By \cref{ass:phi}, we can calculate $\boldsymbol{K}_{\tau}$ and $\boldsymbol{Z}_{\tau}$ without knowing $\phi$.

\begin{assumption}
    Each entry of $\boldsymbol{K}_{\tau}$ and $\boldsymbol{Z}_{\tau}$ can be calculated by \autoref{eq:k_matrix} and \autoref{eq:z_matrix} respectively, where $t, j = 1, 2, ..., \tau$, and $\kappa_{1}$ and $\kappa_{2}$ are two kernel functions.

\begin{small}
    \begin{align}
    \label{eq:k_matrix}
        (\boldsymbol{K}_{\tau})_{t, j} & = \kappa_{1}(\boldsymbol{x}_{t}, \boldsymbol{x}_{j}) \times \kappa_{1} (\boldsymbol{b}_{t} \boldsymbol{Y}, \boldsymbol{b}_{j} \boldsymbol{Y}) \\[10pt]
    \label{eq:z_matrix}
        (\boldsymbol{Z}_{\tau})_{t, j} & =
    \begin{cases}
        \kappa_{2}(\boldsymbol{b}_{t} \boldsymbol{Y}, \boldsymbol{b}_{j} \boldsymbol{Y}) & \boldsymbol{p}_{t} = \boldsymbol{p}_{j} \\
        0 & \boldsymbol{p}_{t} \neq \boldsymbol{p}_{j}
    \end{cases}
    \end{align}
\end{small}

\label{ass:phi}
\end{assumption}

Denoting the above entry values as $k_{tj}$ and $z_{tj}$, then the kernel vectors at time step $\tau$ are $\boldsymbol{k}_{\tau} = (k_{1 \tau}, k_{2 \tau}, ..., k_{\tau, \tau})$ and $\boldsymbol{z}_{\tau} = (z_{1 \tau}, z_{2 \tau}, ..., z_{\tau, \tau})$, and $\boldsymbol{K}_{\tau}$ and $\boldsymbol{Z}_{\tau}$ are the kernel matrices. Based on \cref{ass:phi}, we have \cref{lemma:new_formulation} to approximate the acceptance function.

\vspace{2mm}
\begin{lemma}
    Instead of learning transformation function $\phi$, parameters $\boldsymbol{\Theta}$ and $\boldsymbol{U}$, and then estimating $r_{\tau + 1} (\boldsymbol{b})$ by \autoref{eq:accept eq.}, it is equivalent to iterate \autoref{eq:a_theta} and \autoref{eq:d_u}, then estimate $r_{\tau + 1} (\boldsymbol{b})$ using \autoref{eq:exploit_by_kernel}. Specifically, $\bar{\boldsymbol{k}}_{\tau + 1} = \boldsymbol{k}_{\tau + 1}[1: \tau]$ and $\bar{\boldsymbol{z}}_{\tau + 1} = \boldsymbol{z}_{\tau + 1}[1: \tau]$, which are $\boldsymbol{k}_{\tau + 1}$ and $\boldsymbol{z}_{\tau + 1}$ without their last entries.

\begin{small}
    \begin{align}
        \label{eq:a_theta}
        & \textcolor{red}{\boldsymbol{A}_{\tau} \textup{vec}(\boldsymbol{\Theta})} = \boldsymbol{K}_{\tau} (\boldsymbol{K}_{\tau} + \lambda_{1} \boldsymbol{I}_{\tau})^{-1} (\boldsymbol{r}_{\tau} - \textcolor{blue}{\boldsymbol{D}_{\tau} \textup{vec}(\boldsymbol{U})}) \\[10pt]
        \label{eq:d_u}
        & \textcolor{blue}{\boldsymbol{D}_{\tau} \textup{vec}(\boldsymbol{U})} = \boldsymbol{Z}_{\tau} (\boldsymbol{Z}_{\tau} + \lambda_{2} \boldsymbol{I}_{\tau})^{-1} (\boldsymbol{r}_{\tau} - \textcolor{red}{\boldsymbol{A}_{\tau} \textup{vec}(\boldsymbol{\Theta})}) \\[10pt]
        \label{eq:exploit_by_kernel}
        & \bar{r}_{\tau + 1} (\boldsymbol{b}) = \bar{\boldsymbol{k}}_{\tau + 1} (\boldsymbol{K}_{\tau} + \lambda_{1} \boldsymbol{I}_{\tau})^{-1} (\boldsymbol{r}_{\tau} - \textcolor{blue}{\boldsymbol{D}_{\tau} \textup{vec}(\boldsymbol{U})}) \\ 
        & \phantom{\bar{r}_{\tau + 1} (\boldsymbol{b}) = } + \bar{\boldsymbol{z}}_{\tau + 1} (\boldsymbol{Z}_{\tau} + \lambda_{2} \boldsymbol{I}_{\tau})^{-1} (\boldsymbol{r}_{\tau} - \textcolor{red}{\boldsymbol{A}_{\tau} \textup{vec}(\boldsymbol{\Theta})}) \notag
\end{align}
\end{small}

\label{lemma:new_formulation}
\end{lemma}

\vspace{-2pt}
Considering the definitions of $\boldsymbol{A}_{\tau}$ and $\boldsymbol{D}_{\tau}$, it is evident that $\boldsymbol{A}_{\tau} \text{vec}(\boldsymbol{\Theta})$ and $\boldsymbol{D}_{\tau} \text{vec}(\boldsymbol{U})$ are partial acceptances corresponding to the two terms in \autoref{eq:accept eq.} for historical time steps $t=1, 2, ..., \tau$. From the iteration results, the two terms in \autoref{eq:exploit_by_kernel} estimate the respective terms in \autoref{eq:accept eq.} for the subsequent time step $\tau + 1$.

\subsubsection{NegUCB Algorithm}
\label{subsubsec:NegUCB_alg}
For the subsequent time step $\tau +1$, we can estimate $r_{\tau + 1}(\boldsymbol{b})$ for each bid $\boldsymbol{b} \in B_{\tau + 1}$ using \autoref{eq:exploit_by_kernel}, then choose a bid to put forth or decide to accept or reject the bid from the counterpart by \autoref{eq:bid eq.}. However, this approach relies solely on exploiting historical data, which may lead to sub-optimal choices. Hence, we explore the estimation uncertainty based on the Upper Confidence Bound principle~\cite{LinUCB, KernelUCB, TCB}. 

Instead of \autoref{eq:bid eq.}, we make decisions by \autoref{eq:explore}, where $e_{\tau +1}$ measures the estimation variance and is expressed in \autoref{eq:ucb_kernel}. Parameters $\alpha_{\theta}$ and $\alpha_{u}$ are elaborated in \cref{lemma:bound}. For notation conciseness, we use $k_{\tau + 1}$ and $z_{\tau + 1}$ to denote $k_{\tau + 1, \tau + 1}$ and $z_{\tau + 1, \tau + 1}$.

\vspace{-5pt}
\begin{small}
    \begin{align}
        \label{eq:explore}
        & \boldsymbol{b}_{\tau + 1}^{*} = \mathop{\arg\max} \left \{ \bar{r}_{\tau+1} (\boldsymbol{b}) + e_{\tau+1} (\boldsymbol{b}) \right \} \times f_{\tau +1} (\boldsymbol{b}) \\[10pt]
        \label{eq:ucb_kernel}
        & e_{\tau + 1}(\boldsymbol{b}) = \frac{\alpha_{\theta}}{\sqrt{\lambda_{1}}} \sqrt{k_{\tau + 1} - \bar{\boldsymbol{k}}_{\tau + 1} (\boldsymbol{K}_{\tau} + \lambda_{1} \boldsymbol{I}_{\tau})^{-1} \bar{\boldsymbol{k}}_{\tau + 1}^{\mathsf{T}}} \\
        & \phantom{e_{\tau + 1}(\boldsymbol{b}) = } + \frac{\alpha_{u}}{\sqrt{\lambda_{2}}} \sqrt{z_{\tau + 1} - \bar{\boldsymbol{z}}_{\tau + 1} (\boldsymbol{Z}_{\tau} + \lambda_{2} \boldsymbol{I}_{\tau})^{-1} \bar{\boldsymbol{z}}_{\tau + 1}^{\mathsf{T}}} \notag
    \end{align}
\end{small}

Integrating exploitation and exploration, NegUCB is implemented online as \cref{alg:algorithm}, where we use $\boldsymbol{a}_{\tau}$ and $\boldsymbol{d}_{\tau}$ to respectively denote $\boldsymbol{A}_{\tau} \text{vec}(\boldsymbol{\Theta})$ and $\boldsymbol{D}_{\tau} \text{vec}(\boldsymbol{U})$ for notation conciseness. \textit{Online} means the parameters are incrementally updated each time new negotiation data is generated. NegUCB essentially iterates between \textbf{Step 1.} Estimating the second term in \autoref{eq:accept eq.}, then calculating the first term; \textbf{Step 2.} Estimating the first term in \autoref{eq:accept eq.}, then calculating the second term.

\begin{algorithm}[tb]
   \caption{NegUCB Algorithm}
   \label{alg:algorithm}
\begin{algorithmic}
   \STATE {\bfseries Input:} $\lambda_{1}, \lambda_{2} \in (0, + \infty)$, kernel functions $\kappa_{1}$, $\kappa_{2}$
   \STATE{\bfseries Output:} vectors $\textcolor{red}{\boldsymbol{a}_{N}}$ and $\textcolor{blue}{\boldsymbol{d}_{N}}$
   
   \FOR{$\tau=1$ {\bfseries to} $N$}
   \STATE select bid $\boldsymbol{b}_{\tau}$ randomly if $\tau=1$, or according to \autoref{eq:explore} if $\tau > 1$, and observe $r_{\tau}$
   \IF{$\tau = 1$}
   \STATE initialize $\textcolor{blue}{\boldsymbol{d}_{\tau - 1}}$ as an empty vector
   \STATE initialize kernel matrix $\boldsymbol{Z}_{\tau } = [z_{\tau}]$ and set $a_{\tau} = r_{\tau}$
   \ELSE
   \STATE kernel matrix $\boldsymbol{Z}_{\tau} = [\boldsymbol{Z}_{\tau - 1}, \bar{\boldsymbol{z}}_{\tau}^{\mathsf{T}}; \bar{\boldsymbol{z}}_{\tau}, z_{\tau}]$
   \STATE calculate $a_{\tau} = r_{\tau} - \bar{\boldsymbol{z}}_{\tau} (\boldsymbol{Z}_{\tau - 1} + \lambda_{2} \boldsymbol{I}_{\tau-1})^{-1} \textcolor{blue}{\boldsymbol{d}_{\tau - 1}}$
   \ENDIF
   
   \IF{$\tau = 1$}
   \STATE initialize kernel matrix $\boldsymbol{K}_{\tau } = [k_{\tau}]$ and $\textcolor{red}{\boldsymbol{a}_{\tau}} = (a_{\tau})$
   \ELSE 
   \STATE kernel matrix $\boldsymbol{K}_{\tau} = [\boldsymbol{K}_{\tau - 1}, \bar{\boldsymbol{k}}_{\tau}^{\mathsf{T}}; \bar{\boldsymbol{k}}_{\tau}, k_{\tau}]$
   \STATE $\textcolor{red}{\boldsymbol{a}_{\tau}} = (\textcolor{red}{\boldsymbol{a}_{\tau -1}}; a_{\tau})$
   \ENDIF

   \STATE calculate $d_{\tau} = r_{\tau} - \boldsymbol{k}_{\tau} (\boldsymbol{K}_{\tau} + \lambda_{1} \boldsymbol{I}_{\tau})^{-1} \textcolor{red}{\boldsymbol{a}_{\tau}}$
   \STATE $\textcolor{blue}{\boldsymbol{d}_{\tau}} = (\textcolor{blue}{\boldsymbol{d}_{\tau -1}}; d_{\tau})$
   \ENDFOR

\end{algorithmic}
\end{algorithm}

\subsection{Theoretical Analysis to NegUCB}
\label{sec:theoretical_analysis}

\begin{figure*}[t!]
    \centering
    \includegraphics[width=0.9\textwidth, trim={4cm 0 4cm 1cm}, clip]{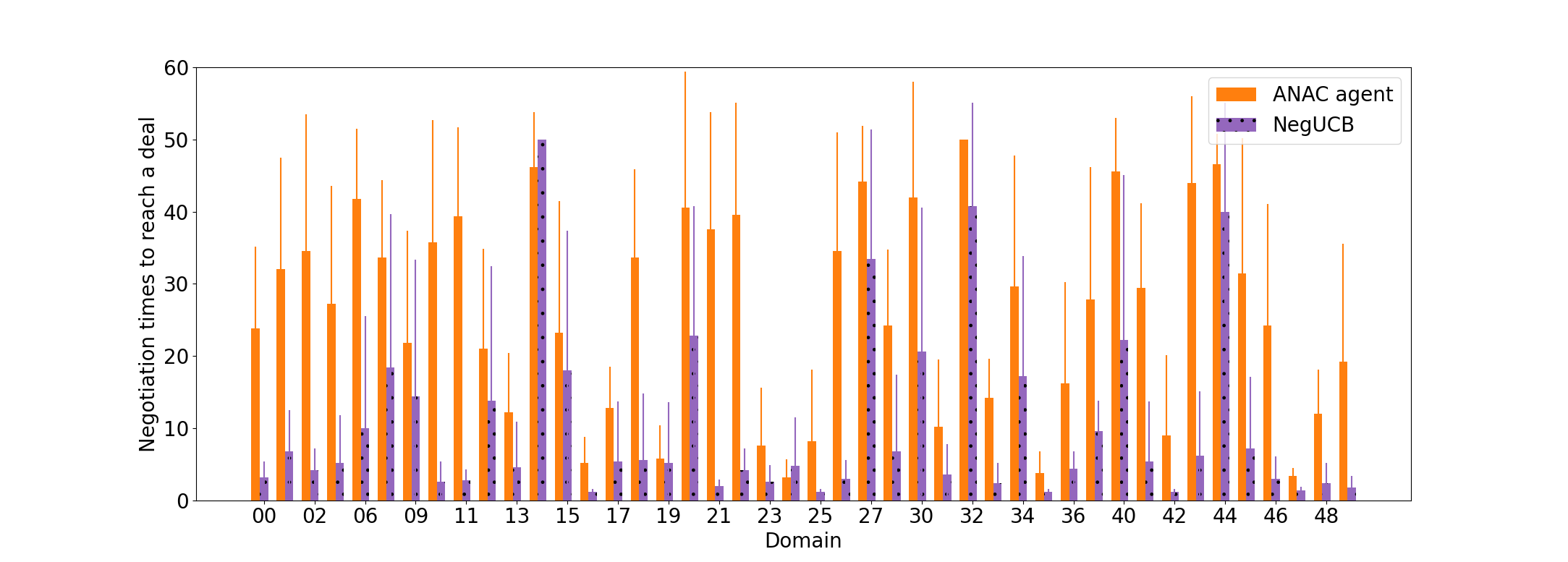}
    \\[-10pt]
    \caption{Negotiation steps needed to reach a deal on each ANAC domain of \textit{domain 00 - 49}.}
    \label{fig:anac_neg_times}
    \vspace{-10pt}
\end{figure*}

\begin{lemma}
    If the true parameters satisfy $\left \| \boldsymbol{\Theta}_{*} \right \| \leq \beta_{\theta}$ and $\left \| \boldsymbol{U}_{*} \right \| \leq \beta_{u}$, the samples satisfy $\left \| \phi(\boldsymbol{b}_{t} \boldsymbol{Y}) \otimes \phi(\boldsymbol{x}_{t}) \right \| \leq 1$ and $\left \| \phi(\boldsymbol{b}_{t} \boldsymbol{Y}) \otimes \boldsymbol{p}_{t} \right \| \leq 1$ for $t=1, 2, ..., \tau$, then with probability at least $1-\sqrt{\delta}$, the two terms in \autoref{eq:exploit_by_kernel} have estimation error bounds $\alpha_{\theta}$ and $\alpha_{u}$ as follows. Here $h_{*}$ and $m_{*}$ are the effective dimensions of $\boldsymbol{\mathcal{A}}_{\tau} = \boldsymbol{A}_{\tau}^{\mathsf{T}} \boldsymbol{A}_{\tau} + \lambda_{1} \boldsymbol{I}_{h^{2}}$ and $\boldsymbol{\mathcal{D}}_{\tau} = \boldsymbol{D}_{\tau}^{\mathsf{T}} \boldsymbol{D}_{\tau} + \lambda_{2} \boldsymbol{I}_{mh}$, $p, q \in (0, 1)$ are constants.

\begin{small}
    \begin{align}
        \alpha_{\theta} = & \left \| \textup{vec}(\boldsymbol{\Theta}_{\tau}) - \textup{vec}(\boldsymbol{\Theta}_{*}) \right \|_{\boldsymbol{\mathcal{A}}_{\tau}} \\ \notag
        \leq & \lambda_{1} \beta_{\theta} + \sqrt{h_{*} \textup{log} (1 + \frac{\tau}{\lambda_{1} h_{*}}) - \textup{log} \delta} + \frac{2 \beta_{u}}{\sqrt{\lambda_{1}} q} \\[10pt]
        \alpha_{u} = & \left \| \textup{vec}(\boldsymbol{U}_{\tau}) - \textup{vec}(\boldsymbol{U}_{*}) \right \|_{\boldsymbol{\mathcal{D}}_{\tau}} \\ \notag
        \leq & \lambda_{2} \beta_{u} + \sqrt{m_{*} \textup{log} (1 + \frac{\tau}{\lambda_{2} m_{*}}) - \textup{log} \delta} + \frac{2 \beta_{\theta}}{\sqrt{\lambda_{2}} p}
    \label{eq:bound}
\end{align}
\end{small}

\label{lemma:bound}
\end{lemma}

\vspace{2mm}
In \cref{lemma:bound}, $\boldsymbol{\mathcal{A}}_{\tau}$ and $\boldsymbol{\mathcal{D}}_{\tau}$ correspond to the first item of \autoref{eq:theta} and \autoref{eq:u}, respectively. \textit{Effective dimension}~\cite{KernelUCB, Gaussian} is a commonly used concept in kernel regression and can be considered as the number of principal dimensions. They contract the bounds as $h_{*} \ll h^{2}$ and $m_{*} \ll mh$, where $h$ is the dimension of $\mathcal{H}$. Bounds of each sample $\phi(\boldsymbol{b}_{t} \boldsymbol{Y}) \otimes \phi(\boldsymbol{x}_{t})$ and $\phi(\boldsymbol{b}_{t} \boldsymbol{Y}) \otimes \boldsymbol{p}_{t}$ are set as $1$ for description convenience. They correlate with the number of items in the bid, referred to as the \textit{bid cardinality} and denoted as $\gamma \leqslant n \in \mathbb{Z}^{+}$. We can guarantee the bounds of samples by normalizing the contexts $\boldsymbol{X}$ and $\boldsymbol{Y}$. Based on \cref{lemma:bound}, we guarantee the performance of NegUCB by the following theorem.

\vspace{5pt}
\begin{theorem}
\label{theorem:regret}
    Under the same assumptions as \cref{lemma:bound}, with probability at least $1-\sqrt{\delta}$, the cumulative regret of \cref{alg:algorithm} has the following upper bound, where $r_{t} (\boldsymbol{b}_{t}^{*})$ and $r_{t} (\boldsymbol{b}_{t})$ are respectively the true acceptance of the optimal bid $\boldsymbol{b}_{t}^{*}$ and the bid chosen by \cref{alg:algorithm} at time step $t$. $\alpha_{f}$ is the union bound of the benefit functions, i.e., $|f_{t} (\boldsymbol{b})| \leq \alpha_{f}$ for $\forall \boldsymbol{b} \in B_{t}$ and $\forall t \in \left \{ 1, 2, ..., \tau \right \}$.

\begin{small}
    \begin{align}
    \begin{split}
        & \sum_{t=0}^{\tau} r_{t} (\boldsymbol{b}_{t}^{*}) \times f_{t} (\boldsymbol{b}_{t}^{*}) - r_{t} (\boldsymbol{b}_{t}) \times f_{t} (\boldsymbol{b}_{t}) \\ 
        \leq & 2 \alpha_{\theta} \alpha_{f} \sqrt{2 h_{*} \tau \textup{log} (1 + \frac{\tau}{\lambda_{1} h_{*}})} \\ 
        & + 2 \alpha_{u} \alpha_{f} \sqrt{2 m_{*} \tau \log (1 + \frac{\tau}{\lambda_{2} m_{*}})}
    \end{split}
\end{align}
\end{small}

    \vspace{-5pt}
\end{theorem}

Indeed, the cumulative regret is sub-linear concerning the number of time steps $\tau$. This implies that as the number of negotiation steps increases, the cumulative regret grows at a slower rate, indicating improved negotiation capability. Besides, the bound is independent of the bid cardinality $\gamma$, distinguishing NegUCB from existing works~\cite{C2UCB, ComLinUCB, CC-MAB, ETCG, RGL}. It is a result of the full-bandit feedback and \cref{ass:psi}. The effect from bid cardinality to the cumulative regret bound is further discussed in \autoref{app:cumulative_regret_analysis}.

\section{Experiments}
\label{sec:experiment}

In this section, we evaluate NegUCB across the three representative negotiation tasks depicted in \cref{fig:neg_example}, comparing it with five representative baselines: ANAC agent \footnote{\url{http://ii.tudelft.nl/nego/node/7}}, LinUCB, FactorUCB, KernelUCB, and a reinforcement learning-based negotiation method \cite{Kris_Cao, Anegma}. It is important to note that we extend the original UCB-based baselines to handle combinatorial bandits and full-bandit feedback effectively. Further analysis regarding the rationale behind baseline selection is in \autoref{app:baseline_seletion}.

As mentioned, the benefit function $f_{\tau}$ is problem-specific. In our experiments, we set $f_{\tau} (\boldsymbol{b}_{\tau}) = 1$ if $\boldsymbol{b}_{\tau} \in \mathcal{C} \cap B_{\tau}$, where $\mathcal{C}$ consists of bids satisfying certain beneficial constraints, otherwise, $f_{\tau} (\boldsymbol{b}_{\tau}) = 0$. It implies that we encourage bids that are advantageous to us. This simple configuration lets us concentrate on the acceptance function $r_{\tau}$ rather than the handcrafted $f_{\tau}$. The subsections of specific tasks will further define the set $\mathcal{C}$ constraining bids.

\begin{figure*}[t!]
     \centering
     \subfigure[$\times$ 1000: Cumulative theoretical regret]{
     \includegraphics[width=0.31\linewidth, trim={0 0 1.5cm 1.5cm}, clip]{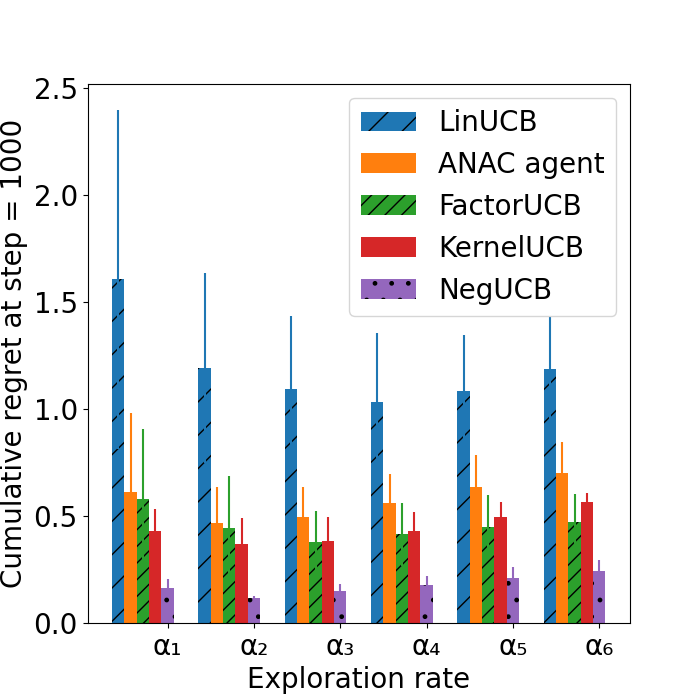}
     \label{fig:syn_1}
     }
     \hfill
     \subfigure[$\times$ 100: Cumulative theoretical regret]{
     \includegraphics[width=0.31\linewidth, trim={0 0 1.5cm 1.5cm}, clip]{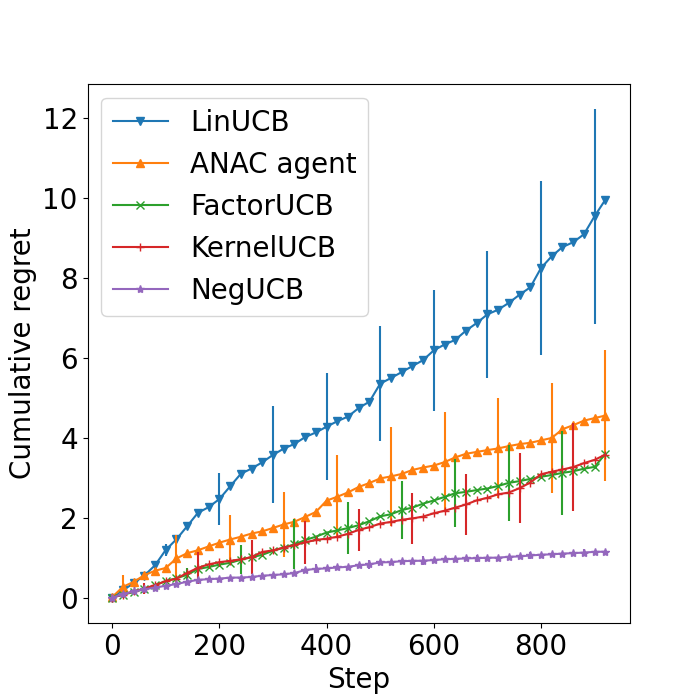}
     \label{fig:syn_2}
     }
     \hfill
     \subfigure[$\times$ 100: Cumulative acceptance regret]{
     \includegraphics[width=0.31\linewidth, trim={0 0 1.5cm 1.5cm}, clip]{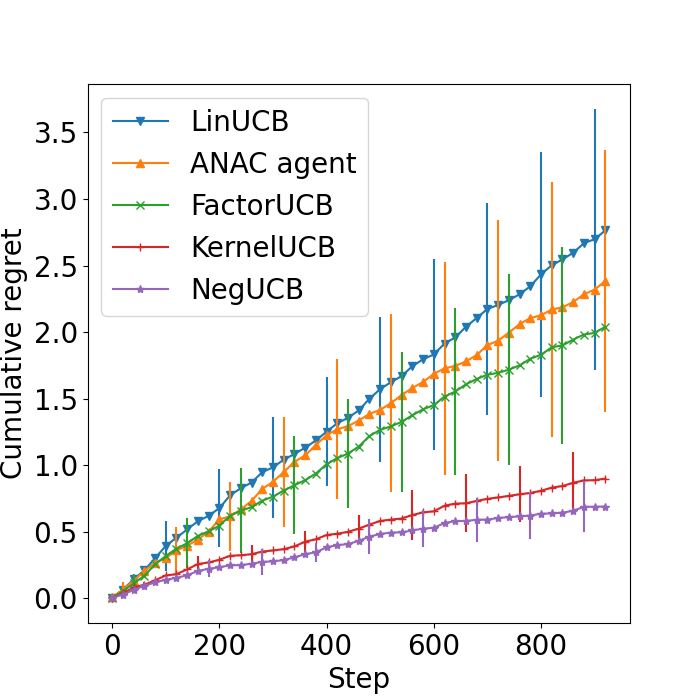}
     \label{fig:syn_3}
     }
     \\[-5pt]
    \caption{Experiment results of resource allocation task. \textit{Theoretical regret} represents the difference between the estimated $\bar{r}$ and the simulated $r$. \textit{Acceptance regret} refers to the difference between the estimated and simulated acceptance.}
    \label{fig:syn_exp}
    \vspace{-5pt}
\end{figure*}

\begin{figure}[t!]
     \centering
        \includegraphics[width=0.4\textwidth, trim={0 0 1cm 1.5cm}, clip]{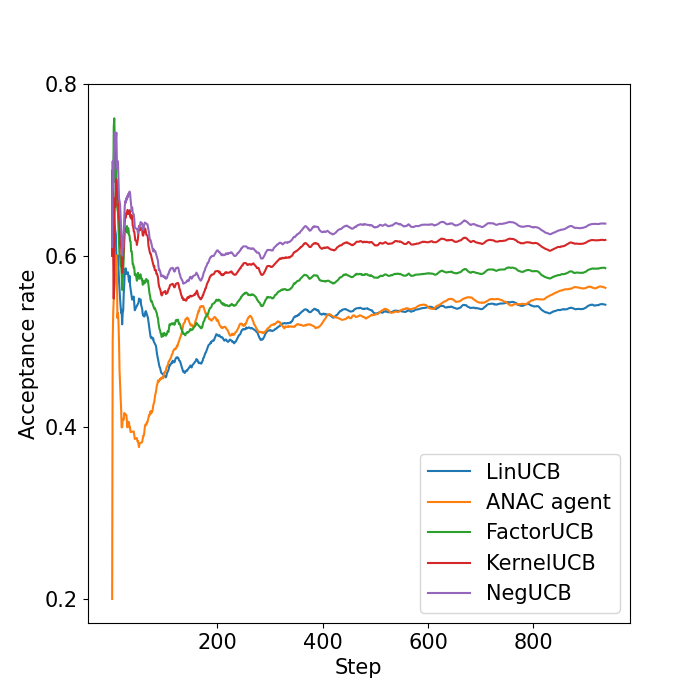}
    \vspace{-10pt}
     \caption{Acceptance rate on resource allocation task, which is defined as the percentage of the proposed bids being accepted.}
     \label{fig:syn_acceptance_rate}
    \vspace{-10pt}
\end{figure}

\subsection{Multi-issue Negotiation}

ANAC (Automated Negotiating Agents Competition) is an international tournament that has been held since 2010, providing 50 negotiation domains. However, compared to the settings of NegUCB, ANAC tasks are relatively simple. For instance, negotiators and items lack contexts, and there is only one negotiator pair for each domain. Consequently, some components of NegUCB are not necessary for these tasks. In this subsection, we modify NegUCB for compatibility with ANAC tasks, showing the adaptability of NegUCB to diverse negotiation problems. In ANAC experiment, NegUCB does not consider any context and relies on inferring hidden states of negotiators and items from negotiation experiences. Essentially, it degenerates into traditional combinatorial bandits. In contrast, most of the ANAC agents submitted by tournament participants, including the winners~\cite{ANAC}, are rule-based. 

Original ANAC tasks impose a strict deadline on negotiation, limiting each negotiation pair to a constant number of negotiation steps. Negotiators are aware of this deadline and can strategically utilize it. This setup diverges from our setting in that a negotiator may lose patience at any time, and its counterpart may not be aware of it. Therefore, in this subsection, we redefine the task. First, we eliminate the deadline and investigate the number of rounds needed to reach a deal. Second, we define the constraining set $\mathcal{C}$ only contains bids whose utilities are larger than the mean utility of all possible bids to our negotiator. Agents achieving a deal in fewer steps are more effective. Based on the insights of the ANAC agents submitted by tournament participants, we modify them to be compatible with the redefined task. Specifically, the ANAC agent we adopt in this experiment randomly selects a valid bid from those with the highest utility rankings for our negotiator.

We investigate the number of negotiation steps required to achieve a deal by each algorithm across $50$ ANAC domains, specifically from \textit{domain 00} to \textit{domain 49}. For \textit{domain 3}, \textit{domain 4}, \textit{domain 7}, \textit{domain 28}, \textit{domain 37}, and \textit{domain 38}, both algorithms failed to reach a deal in $50$ rounds, thus for the sake of conciseness, we show the results on the remaining $44$ domains in \cref{fig:anac_neg_times}. NegUCB consistently achieves beneficial deals much earlier across almost all ANAC domains. Considering the effectiveness of the simplified NegUCB used in this subsection, it is adopted in place of the ANAC agent in the following experiments for a more appropriate comparison.

Additionally, we analyze the action spaces. Considering \textit{domain 13} for example, it has $4$ issues, each of which has $6, 12, 5, 26$ possible values to choose from, then the bid set contains at most $6 \times 12 \times 5 \times 26 = 9360$ choices. Similarly, other domains exhibit comparable action space sizes.

\subsection{Resource Allocation}
Motivated by experiments of existing works~\cite{Kris_Cao}, we design a resource allocation task. 

Assume there are three categories of items, and the number of items in each category does not exceed $5$. Each item category has a randomly generated context vector denoted as $\boldsymbol{y}_{j}, j=1, 2, 3$. A context vector $\boldsymbol{x}_{w}$ and a hidden state vector $\boldsymbol{u}_{w}$ are randomly generated for each of the $30$ negotiator pairs. For simplicity, we assume that $\boldsymbol{x}_{w}$, $\boldsymbol{y}_{j}$, and $\boldsymbol{u}_{w}$ are all 2-dimensional, with each entry in the range $[0, 1]$. The acceptance function is simulated using \autoref{eq:accept eq.}, where the transformation function is as \autoref{eq:transforming_func}. Besides, we draw the parameter matrix $\boldsymbol{\Theta}$ of size $6 \times 6$ from a Gaussian distribution $\mathcal{N}(0, 1)$. The counterpart accepts the bid if the simulated acceptance $r$ satisfies $r > 0$. 



\vspace{-5pt}
\begin{small}
    \begin{equation}
    \phi(\boldsymbol{x}) = ( \frac{1}{\sqrt{2}}, x_{1}, x_{2}, \frac{1}{\sqrt{2}} x_{1}^{2}, x_{1} x_{2}, \frac{1}{\sqrt{2}} x_{2}^{2} )
    \label{eq:transforming_func}
\end{equation}
\end{small}

\noindent Specifically, the transformation function is the basis function of polynomial kernel $\kappa(\boldsymbol{x}_{w}, \boldsymbol{x}_{j}) = \frac{1}{2} (\boldsymbol{x}_{w} \boldsymbol{x}_{j}^{\mathsf{T}} + 1)^{2}$. Furthermore, we define the set $\mathcal{C}$ contains bids that allow our negotiator to acquire more items than the counterpart.

\cref{fig:syn_1} shows the cumulative theoretical regret for each algorithm under various exploration parameters, i.e., $\alpha_{\theta} = \alpha_{u} = \alpha_{1}, \alpha_{2}, ..., \alpha_{6}$ summarized in \autoref{app:resource_allocation}. It is evident that the cumulative theoretical regret for each algorithm decreases initially and then increases, illustrating the advantages of exploration and the drawbacks of over-exploration. \cref{fig:syn_2} and \cref{fig:syn_3} display the cumulative theoretical regret and cumulative acceptance regret of each algorithm at each time step under their corresponding optimal exploration parameter, respectively. \cref{fig:syn_acceptance_rate} illustrates the acceptance rate of each algorithm under their corresponding optimal exploration parameter. Given the random nature of exploration in reinforcement learning, i.e., $\epsilon$-greedy, we extend the training duration of the reinforcement learning method to $20000$ steps to ensure the results accurately reflect its true capabilities. Its final result reaches an acceptance rate lower than $0.6$. Refer to \autoref{app:resource_allocation} for a detailed insight into its convergence process. From these results, we can observe clear advantages of NegUCB.

In this experiment, the action space comprises at most $6 \times 6 \times 6 = 216$ actions, which aligns with that in the existing work \cite{Kris_Cao} to confirm the effectiveness of NegUCB for established problems. We add another experiment with a larger action space in \cref{app:resource_allocation}.

\subsection{Trading}

CivRealm~\cite{CivRealm} is an interactive environment designed for the open-source strategy game \textit{Freeciv}. In this environment, multiple players engage in their civilizations' simultaneous development and competition. Alongside elements such as land, population, and economy, each player possesses a technology tree, allowing them to research and acquire the $87$ technologies progressively.

\begin{figure}[H]
     \centering
        \includegraphics[width=0.37\textwidth, trim={0 0 1cm 1.7cm}, clip]{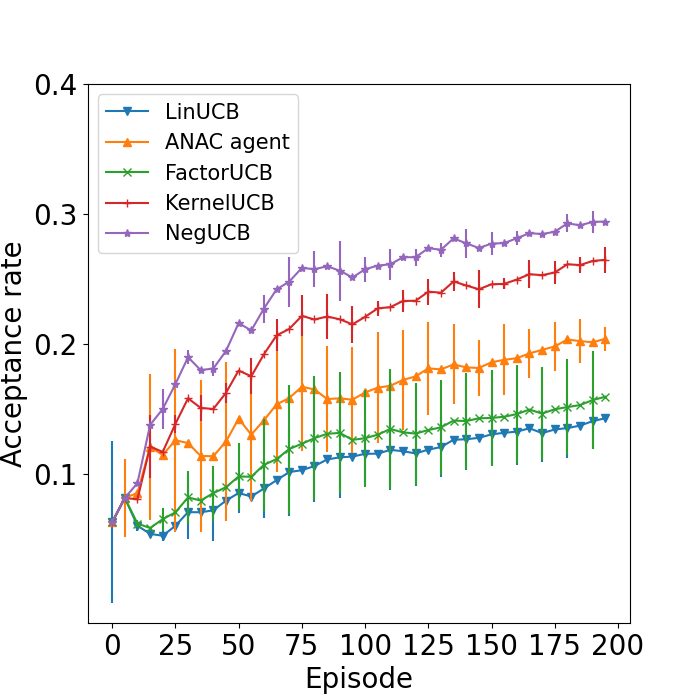}
    \vspace{-5pt}
     \caption{Acceptance rate at each episode on trading task}
     \vspace{-15pt}
     \label{fig:civ_2}
\end{figure}

\begin{figure}[H]
    \centering
    \includegraphics[width=0.4\textwidth]{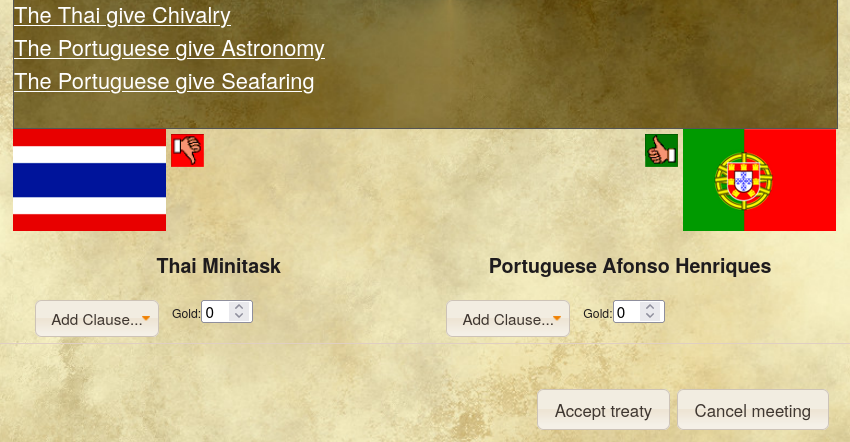}
    \caption{A case of negotiation in CivRealm. Thailand is our negotiator, while Portugal is our counterpart.}
    \label{fig:neg_case}
    \vspace{-10pt}
\end{figure}

One crucial feature of CivRealm is its \textit{Diplomacy} component, enabling players to engage in technology trades. For instance, if our negotiator possesses the technology \textit{Chivalry} and seeks the technology \textit{Astronomy}, besides researching it by itself, our negotiator can also acquire it through trading with other players who already possess \textit{Astronomy}. A negotiation window of CivRealm is as \cref{fig:neg_case} \footnote{It is from a running Freeciv game and may contain politically sensitive names of nations, which are purely hypothetical.}. A negotiator can counter-propose or cancel the meeting if they reject the bid. On the other hand, the negotiator can accept the bid by accepting the treaty. In this experiment, $\boldsymbol{b}_{\tau} \in \mathcal{C}$ if the total cost of the given technologies is no more than that of the required ones. For practical reasons, we set the bid cardinality as $\gamma = 4$, with details explained in \autoref{app:trading}.

In this experiment, we systematically explore SE kernels with diverse hyper-parameters $\sigma$ to fine-tune the most suitable kernel function for the technology trading task in CivRealm. Based on the results, we conclude that the SE kernel with $\sigma=1$ emerges as the most suitable choice for this task. Besides, we tune the exploration rate ranging from $0$ to $1$ and choose $0.1$ as the optimal exploration rate for NegUCB. Please refer to \autoref{app:trading} for more details. Surprisingly, apart from the baselines LinUCB, KernelUCB, and our proposed method NegUCB, other baselines fail to demonstrate improvements with increased exploration. We attribute this observation to the complexity of the task, where inaccurate formulations result in misguided exploration strategies. \cref{fig:civ_2} illustrates the acceptance rates of each algorithm under their corresponding optimal exploration parameters, i.e., $0.1, 0, 0, 0.1, 0.1$, affirming the clear advantages of NegUCB. The reinforcement learning method is not utilized in this experiment due to the challenge associated with handling such a large action space, whose cardinality is at most $\sum_{j=1}^{\gamma} \binom{87}{j}$.

\textbf{Case Study.} A negotiation case on CivRealm is depicted in \cref{fig:neg_case}. Thailand proposed to give \textit{Chivalry} and seek \textit{Astronomy} and \textit{Seafaring} from Portugal with costs of $270$, $185$, and $112$, respectively. The net income for Portugal would be $270 - 185 - 112 = -27$. However, according to the running game, Portugal accepted the bid. It suggests the presence of hidden states that we did not observe, influencing Portugal's decision to accept the bid. Without the hidden state component in NegUCB, we might overlook such bids, substantiating that hidden states play a crucial role in estimating the counterpart's decisions.

\section{Conclusion}
\label{sec:conclusion}

This paper introduces a comprehensive formulation for negotiation, grounded in contextual combinatorial multi-armed bandits, capable of encompassing a broad spectrum of real-world negotiation tasks. Building upon this formulation, we propose the NegUCB algorithm as a solution to address the four prevalent challenges in negotiation: the exploitation-exploration dilemma, handling large action spaces, partial observations, and complex acceptance functions. Under mild assumptions, NegUCB ensures a regret upper bound that is sub-linear with respect to the negotiation steps and independent of the bid cardinality. A series of experiments on diverse negotiation tasks validate NegUCB's effectiveness and advantages in learning negotiation strategies.

\section*{Acknowledgements}
This work was supported by the National Natural Science Foundation of China.


\section*{Impact Statement}
This paper aims to advance the field of negotiation among agents with diverse interests from a multi-armed bandit perspective, which is well-suited for negotiation problems and has been well-investigated. It encourages future research to tackle any limitation of our method under the general formulation utilizing the advantages of bandit-based techniques. The model applies to scenarios where the agent knows the negotiation target it is trying to reach with the counterpart at each time step, which is a very mild constraint in \textit{negotiation}. However, for tasks where hypermetropic \textit{planning} plays the main role and negotiation is only one type of action that merely assists the agent in completing the task, i.e., those fall in the reinforcement learning paradigm, the model presented in this work may be limited due to the lack of an explicitly given negotiation target.

\bibliography{example_paper}
\bibliographystyle{icml2024}

\newpage
\appendix
\onecolumn

\section{Proofs}
In this section, we first provide a derivation of the closed-form solutions in \cref{app:closed_form_solution}, then we provide the proofs of \cref{lemma:new_formulation}, \cref{lemma:bound} and \cref{theorem:regret} in \cref{app:lemma_3_3}, \cref{app:lemma_3_4}, and \cref{app:theorem_3_5}, respectively.

\subsection{Derivation of Closed-form Solutions}
\label{app:closed_form_solution}
Under \cref{ass:psi}, the approximated acceptance function and objective function are respectively:

\begin{displaymath}
\begin{split}
        & \bar{r}_{\tau} (\boldsymbol{b}_{\tau}) = \phi(\boldsymbol{x}_{\tau}) \boldsymbol{\Theta} \phi(\boldsymbol{Y}^{\mathsf{T}} \boldsymbol{b}_{\tau}^{\mathsf{T}}) + \boldsymbol{p}_{\tau} \boldsymbol{U} \phi(\boldsymbol{Y}^{\mathsf{T}} \boldsymbol{b}_{\tau}^{\mathsf{T}}) \\
        & \mathcal{L} = \sum_{t=1}^{\tau} \left | \phi(\boldsymbol{x}_{t}) \boldsymbol{\Theta} \phi(\boldsymbol{Y}^{\mathsf{T}} \boldsymbol{b}_{t}^{\mathsf{T}}) + \boldsymbol{p}_{t} \boldsymbol{U} \phi(\boldsymbol{Y}^{\mathsf{T}} \boldsymbol{b}_{t}^{\mathsf{T}}) - r_{t} \right |^{2} + \lambda_{1} \left \| \boldsymbol{\Theta} \right \|^{2} + \lambda_{2} \left \| \boldsymbol{U} \right \|^{2}
\end{split}
\end{displaymath}

Based on basic linear algebra, we can derive the closed-form solutions of $\boldsymbol{\Theta}$ and $\boldsymbol{U}$ easily~\cite{LinUCB, FactorUCB}. The core method we employ in the derivation relies on the conclusion that $\boldsymbol{a} \boldsymbol{B} \boldsymbol{c}^{\mathsf{T}} = (\boldsymbol{c} \otimes \boldsymbol{a}) \text{vec} (\boldsymbol{B})$, where $\boldsymbol{a}$ and $\boldsymbol{c}$ denote any two row-vectors, and $\boldsymbol{B}$ denotes any matrix, provided that their sizes match.

\subsection{Proof of \cref{lemma:new_formulation}}
\label{app:lemma_3_3}

Proof in this subsection directly uses the closed-form solutions in \autoref{eq:theta} and \autoref{eq:u}.
\begin{proof}

According to the closed-form solution in \autoref{eq:theta}, we have the following equation.

\begin{displaymath}
\begin{split}
    & (\boldsymbol{A}_{\tau}^{\mathsf{T}} \boldsymbol{A}_{\tau} + \lambda_{1} \boldsymbol{I}_{h^{2}}) \text{vec}(\boldsymbol{\Theta }) = \boldsymbol{A}_{\tau}^{\mathsf{T}} (\boldsymbol{r}_{\tau} - \boldsymbol{D}_{\tau} \text{vec}(\boldsymbol{U})) \\
    \Rightarrow & \text{vec}(\boldsymbol{\Theta }) = \frac{1}{\lambda_1} (\boldsymbol{A}_{\tau}^{\mathsf{T}} (\boldsymbol{r}_{\tau} - \boldsymbol{D}_{\tau} \text{vec}(\boldsymbol{U})) - \boldsymbol{A}_{\tau}^{\mathsf{T}} \boldsymbol{A}_{\tau} \text{vec}(\boldsymbol{\Theta })) \\ 
    & \phantom{\text{vec}(\boldsymbol{\Theta })} = \frac{1}{\lambda_1} \boldsymbol{A}_{\tau}^{\mathsf{T}} (\boldsymbol{r}_{\tau} - \boldsymbol{D}_{\tau} \text{vec}(\boldsymbol{U}) - \boldsymbol{A}_{\tau} \text{vec}(\boldsymbol{\Theta }))
\end{split}
\end{displaymath}

Denote $\boldsymbol{\alpha} = \frac{1}{\lambda_1} (\boldsymbol{r}_{\tau} - \boldsymbol{D}_{\tau} \text{vec}(\boldsymbol{U}) - \boldsymbol{A}_{\tau} \text{vec}(\boldsymbol{\Theta }))$, thus there is $\text{vec}(\boldsymbol{\Theta }) = \boldsymbol{A}_{\tau}^{\mathsf{T}} \boldsymbol{\alpha}$. Integrating these two equations:

\begin{displaymath}
    \begin{split}
        \Rightarrow & \boldsymbol{\alpha} = \frac{1}{\lambda_1} (\boldsymbol{r}_{\tau} - \boldsymbol{D}_{\tau} \text{vec}(\boldsymbol{U}) - \boldsymbol{A}_{\tau} \boldsymbol{A}_{\tau}^{\mathsf{T}} \boldsymbol{\alpha}) \\
        \Rightarrow & \boldsymbol{\alpha} = (\boldsymbol{A}_{\tau} \boldsymbol{A}_{\tau}^{\mathsf{T}} + \lambda_{1} \boldsymbol{I}_{\tau})^{-1} (\boldsymbol{r}_{\tau} - \boldsymbol{D}_{\tau} \text{vec}(\boldsymbol{U})) \\
        \Rightarrow & \text{vec}(\boldsymbol{\Theta }) = \boldsymbol{A}_{\tau}^{\mathsf{T}} \boldsymbol{\alpha} = \boldsymbol{A}_{\tau}^{\mathsf{T}} (\boldsymbol{A}_{\tau} \boldsymbol{A}_{\tau}^{\mathsf{T}} + \lambda_{1} \boldsymbol{I}_{\tau})^{-1} (\boldsymbol{r}_{\tau} - \boldsymbol{D}_{\tau} \text{vec}(\boldsymbol{U}))
    \end{split}
\end{displaymath}

From the definition of rows of matrix $\boldsymbol{A}_{\tau}$, we have:

\begin{displaymath}
\begin{split}
    (\boldsymbol{A}_{\tau} \boldsymbol{A}_{\tau}^{\mathsf{T}})_{tj} = & (\phi(\boldsymbol{b}_{t} \boldsymbol{Y}) \otimes \phi(\boldsymbol{x}_{t})) (\phi(\boldsymbol{b}_{j} \boldsymbol{Y}) \otimes \phi(\boldsymbol{x}_{j}))^{\mathsf{T}} \\ 
    = & (\phi(\boldsymbol{b}_{t} \boldsymbol{Y}) \phi(\boldsymbol{b}_{j} \boldsymbol{Y})^{\mathsf{T}}) \times (\phi(\boldsymbol{x}_{t}) \phi(\boldsymbol{x}_{j}) ^{\mathsf{T}}) \\
    = & \kappa_{1}(\boldsymbol{b}_{t} \boldsymbol{Y}, \boldsymbol{b}_{j} \boldsymbol{Y}) \times \kappa_{1}(\boldsymbol{x}_{t}, \boldsymbol{x}_{j}) = (\boldsymbol{K}_{\tau})_{tj}
\end{split}
\end{displaymath}

The second equality above is from the fact that any row vectors $\boldsymbol{v}_{1}$, $\boldsymbol{v}_{2}$, $\boldsymbol{\nu}_{1}$, $\boldsymbol{\nu}_{2}$ satisfy $(\boldsymbol{v}_{1} \otimes \boldsymbol{\nu}_{1}) (\boldsymbol{v}_{2} \otimes \boldsymbol{\nu}_{2})^{\mathsf{T}} = (\boldsymbol{v}_{1} \boldsymbol{v}_{2}^{\mathsf{T}}) (\boldsymbol{\nu}_{1} \boldsymbol{\nu}_{2}^{\mathsf{T}})$. As a result, we can derive $\boldsymbol{A}_{\tau} \text{vec}(\boldsymbol{\Theta })$ as follows.

\begin{displaymath}
    \begin{split}
        \boldsymbol{A}_{\tau} \text{vec}(\boldsymbol{\Theta }) = & \boldsymbol{A}_{\tau} \boldsymbol{A}_{\tau}^{\mathsf{T}} (\boldsymbol{A}_{\tau} \boldsymbol{A}_{\tau}^{\mathsf{T}} + \lambda_{1} \boldsymbol{I}_{\tau})^{-1} (\boldsymbol{r}_{\tau} - \boldsymbol{D}_{\tau} \text{vec}(\boldsymbol{U})) \\ 
        = & \boldsymbol{K}_{\tau} (\boldsymbol{K}_{\tau} + \lambda_{1} \boldsymbol{I}_{\tau})^{-1} (\boldsymbol{r}_{\tau} - \boldsymbol{D}_{\tau} \textup{vec}(\boldsymbol{U}))
    \end{split}
\end{displaymath}

For the next time step $\tau + 1$, there is:

\begin{displaymath}
\begin{split}
        \phi(\boldsymbol{x}_{\tau + 1}) \boldsymbol{\Theta } \phi(\boldsymbol{Y}^{\mathsf{T}} \boldsymbol{b}_{\tau + 1}^{\mathsf{T}}) = & (\phi(\boldsymbol{b}_{\tau + 1} \boldsymbol{Y}) \otimes \phi(\boldsymbol{x}_{\tau + 1})) \text{vec}(\boldsymbol{\Theta }) \\ 
        = & \bar{\boldsymbol{k}}_{\tau + 1} (\boldsymbol{K}_{\tau} + \lambda_{1} \boldsymbol{I}_{\tau})^{-1} (\boldsymbol{r}_{\tau} - \boldsymbol{D}_{\tau} \textup{vec}(\boldsymbol{U}))
\end{split}
\end{displaymath}

Derivation of $\boldsymbol{D}_{\tau} \text{vec}(\boldsymbol{U})$ is similar, thus we omit it. 

\end{proof}

\subsection{Proof of \cref{lemma:bound}}
\label{app:lemma_3_4}

Denote $\boldsymbol{\mathcal{A}}_{\tau} = \boldsymbol{A}_{\tau}^{\mathsf{T}} \boldsymbol{A}_{\tau} + \lambda_{1} \boldsymbol{I}_{h^{2}}$ and $\boldsymbol{\mathcal{D}}_{\tau} = \boldsymbol{D}_{\tau}^{\mathsf{T}} \boldsymbol{D}_{\tau} + \lambda_{2} \boldsymbol{I}_{mh}$. $\boldsymbol{\Theta}_{*}$ is the true parameter while $\boldsymbol{\Theta}_{\tau}$ is the parameter estimated at time step $\tau$. $\phi(\boldsymbol{b}_{t} \boldsymbol{Y}) \otimes \phi(\boldsymbol{x}_{t})$ is the sample at time step $t=1, 2, ..., \tau$.

\begin{proof}
    The error of the estimated partial acceptance based on contexts corresponding to $\phi(\boldsymbol{b}_{t+1} \boldsymbol{Y}) \otimes \phi(\boldsymbol{x}_{t+1})$ is as follows.

\begin{displaymath}
    \begin{split}
        & \left | \bar{\boldsymbol{k}}_{\tau + 1} (\boldsymbol{K}_{\tau} + \lambda_{1} \boldsymbol{I}_{\tau})^{-1} (\boldsymbol{r}_{\tau} - \boldsymbol{D}_{\tau} \textup{vec}(\boldsymbol{U})) - (\phi(\boldsymbol{b}_{t+1} \boldsymbol{Y}) \otimes \phi(\boldsymbol{x}_{t+1})) \text{vec}(\boldsymbol{\Theta}_{*}) \right | \\
        \leq & \left \| \phi(\boldsymbol{b}_{t+1} \boldsymbol{Y}) \otimes \phi(\boldsymbol{x}_{t+1}) \right \| \left \| \text{vec}(\boldsymbol{\Theta}_{\tau}) - \text{vec}(\boldsymbol{\Theta}_{*}) \right \|_{\boldsymbol{\mathcal{A}}_{\tau}} \\ 
        \leq & \left \| \text{vec}(\boldsymbol{\Theta}_{\tau}) - \text{vec}(\boldsymbol{\Theta}_{*}) \right \|_{\boldsymbol{\mathcal{A}}_{\tau}} \\ 
        = & \left \| \boldsymbol{\mathcal{A}}_{\tau} (\text{vec}(\boldsymbol{\Theta}_{\tau}) - \text{vec}(\boldsymbol{\Theta}_{*})) \right \|_{\boldsymbol{\mathcal{A}}_{\tau}^{-1}} \\ 
        = & \left \| \boldsymbol{A}_{\tau}^{\mathsf{T}} (\boldsymbol{r}_{\tau} - \boldsymbol{D}_{\tau} \text{vec}(\boldsymbol{U}_{\tau -1})) - (\boldsymbol{A}_{\tau}^{\mathsf{T}} \boldsymbol{A}_{\tau} + \lambda_{1} \boldsymbol{I}_{h^{2}}) \text{vec}(\boldsymbol{\Theta}_{*}) \right \|_{\boldsymbol{\mathcal{A}}_{\tau}^{-1}} \\
        = & \left \| \boldsymbol{A}_{\tau}^{\mathsf{T}} (\boldsymbol{r}_{\tau} - \boldsymbol{D}_{\tau} \text{vec}(\boldsymbol{U}_{\tau -1}) - \boldsymbol{A}_{\tau} \text{vec}(\boldsymbol{\Theta}_{*})) - \lambda_{1} \text{vec}(\boldsymbol{\Theta}_{*}) \right \|_{\boldsymbol{\mathcal{A}}_{\tau}^{-1}} \\
        = & \left \| \boldsymbol{A}_{\tau}^{\mathsf{T}} \boldsymbol{D}_{\tau} \text{vec}(\boldsymbol{U}_{*}) - \boldsymbol{A}_{\tau}^{\mathsf{T}} \boldsymbol{D}_{\tau} \text{vec}(\boldsymbol{U}_{\tau -1}) + \boldsymbol{A}_{\tau}^{\mathsf{T}} \boldsymbol{\epsilon}_{\tau} - \lambda_{1} \text{vec}(\boldsymbol{\Theta}_{*}) \right \|_{\boldsymbol{\mathcal{A}}_{\tau}^{-1}} \\
        \leq & \left \| \boldsymbol{A}_{\tau}^{\mathsf{T}} \boldsymbol{D}_{\tau} (\text{vec}(\boldsymbol{U}_{*}) - \text{vec}(\boldsymbol{U}_{\tau -1})) \right \|_{\boldsymbol{\mathcal{A}}_{\tau}^{-1}} + \left \| \boldsymbol{A}_{\tau}^{\mathsf{T}} \boldsymbol{\epsilon}_{\tau} \right \|_{\boldsymbol{\mathcal{A}}_{\tau}^{-1}} + \lambda_{1} \left \| \boldsymbol{\Theta}_{*} \right \|_{\boldsymbol{\mathcal{A}}_{\tau}^{-1}} \\
    \end{split}
\end{displaymath}

The first inequality holds when the minimum eigenvalue of $\boldsymbol{\mathcal{A}}_{\tau}$ is at least $1$. The term $\epsilon_{\tau}$ accounts for sub-Gaussian noise to the acceptance function. Now, consider the first term above:

\begin{displaymath}
    \begin{split}
        & \left \| \boldsymbol{A}_{\tau}^{\mathsf{T}} \boldsymbol{D}_{\tau} (\text{vec}(\boldsymbol{U}_{*}) - \text{vec}(\boldsymbol{U}_{\tau -1})) \right \|_{\boldsymbol{\mathcal{A}}_{\tau}^{-1}} \\ 
        \leq & \frac{1}{\sqrt{\lambda_{1}}} \left \| \boldsymbol{D}_{\tau} (\text{vec}(\boldsymbol{U}_{*}) - \text{vec}(\boldsymbol{U}_{\tau -1})) \right \| \\
        \leq & \frac{1}{\sqrt{\lambda_{1}}} \sum_{t=1}^{\tau} \left \| (\phi(\boldsymbol{b}_{t} \boldsymbol{Y}) \otimes \boldsymbol{p}_{t}) (\text{vec}(\boldsymbol{U}_{*}) - \text{vec}(\boldsymbol{U}_{t-1})) \right \| \\
        \leq & \frac{1}{\sqrt{\lambda_{1}}} \sum_{t=1}^{\tau} \left \|\text{vec}(\boldsymbol{U}_{*}) - \text{vec}(\boldsymbol{U}_{t-1}) \right \| \\
        \leq & \frac{1}{\sqrt{\lambda_{1}}} \sum_{t=1}^{\tau} \left \|\text{vec}(\boldsymbol{U}_{*}) - \text{vec}(\boldsymbol{U}_{0}) \right \| \times q^{t-1} \\
        \leq & \frac{2 \beta_{u}}{\sqrt{\lambda_{1}}} \times \frac{1 - q^{\tau}}{1 - q} \\
        \leq & \frac{2 \beta_{u}}{\sqrt{\lambda_{1}} (1 - q)}
    \end{split}
\end{displaymath}

The second inequality holds because \cref{alg:algorithm} updates parameters \textit{online}. The fourth inequality is based on Uschmajew's work~\cite{hessian, FactorUCB}, that the estimation of $\boldsymbol{U}$ is local $q$-linearly convergent to the optimizer. Specifically, in the above inequations, parameter $q$ satisfies $0<q<1$. For conciseness, we denote $(1-q) \in (0, 1)$ as $q \in (0, 1)$ in \cref{lemma:bound}. Some works~\cite{TCB} simply assume $\left \| \boldsymbol{A}_{\tau}^{\mathsf{T}} \boldsymbol{D}_{\tau} (\text{vec}(\boldsymbol{U}_{*}) - \text{vec}(\boldsymbol{U}_{\tau -1})) \right \|_{\boldsymbol{\mathcal{A}}_{\tau}^{-1}} = 0$ considering $\boldsymbol{U}_{t-1} \rightarrow \boldsymbol{U}_{*}$ when $t \rightarrow \infty$.

For the second term, we leverage the properties of \textit{self-normalized vector-valued martingales}~\cite{linear_bandit}. Assuming $\epsilon_{\tau}$ belongs to a 1-sub-Gaussian process, then with probability at least $1 - \sqrt{\delta}$, there is the following inequality:

\begin{displaymath}
    \begin{split}
        \left \| \boldsymbol{A}_{\tau}^{\mathsf{T}} \boldsymbol{\epsilon}_{\tau} \right \|_{\boldsymbol{\mathcal{A}}_{\tau}^{-1}} \leq \sqrt{\text{log} \frac{\text{det} (\boldsymbol{\mathcal{A}}_{\tau})}{\text{det} (\lambda_{1} \boldsymbol{I}_{h^{2}})} - \text{log} \delta}
    \end{split}
\end{displaymath}

Because of the Determinant-trace inequality, we have:

\begin{displaymath}
\begin{split}
    & \text{det} \bar{(\boldsymbol{\mathcal{A}}_{\tau})} \leq \left ( \frac{\text{trace} (\bar{\boldsymbol{\mathcal{A}}_{\tau}})}{h_{*}} \right )^{h_{*}} \leq \left ( \lambda_{1} + \frac{\tau}{h_{*}} \right )^{h_{*}} \Rightarrow \\ 
    & \text{det} (\boldsymbol{\mathcal{A}}_{\tau}) \approx \text{det} \bar{(\boldsymbol{\mathcal{A}}_{\tau})} \times \lambda_{1}^{h^{2} - h_{*}} \leq \left ( \frac{\text{trace} (\bar{\boldsymbol{\mathcal{A}}_{\tau}})}{h_{*}} \right )^{h_{*}} \times \lambda_{1}^{h^{2} - h_{*}} \leq \left ( \lambda_{1} + \frac{\tau}{h_{*}} \right )^{h_{*}} \times \lambda_{1}^{h^{2} - h_{*}}
\end{split}
\end{displaymath}

In the above inequalities, $\bar{\boldsymbol{\mathcal{A}}_{\tau}}$ denotes the diagonal matrix whose diagonal entries are the eigenvalues of $\boldsymbol{\mathcal{A}}_{\tau}$ corresponding to the \textit{effective dimensions}. It is worth noting that there may be a \textit{small} coefficient on the right side of $\approx$, depending on the definition of the \textit{effective dimension} $h_{*}$. However, we omit this coefficient for the sake of conciseness. Consequently, the second term has the following bound:
\begin{displaymath}
    \begin{split}
        \left \| \boldsymbol{A}_{\tau}^{\mathsf{T}} \boldsymbol{\epsilon}_{\tau} \right \|_{\boldsymbol{\mathcal{A}}_{\tau}^{-1}} \leq \sqrt{\text{log} \frac{\text{det} (\boldsymbol{\mathcal{A}}_{\tau})}{\lambda_{1}^{h_{*}}} - \text{log} \delta} \leq \sqrt{h_{*} \text{log} (1 + \frac{\tau}{\lambda_{1} h_{*}}) - \text{log} \delta}
    \end{split}
\end{displaymath}

According to the assumptions in \cref{lemma:bound}, we have:

\begin{displaymath}
    \lambda_{1} \left \| \boldsymbol{\Theta}_{*} \right \|_{\boldsymbol{\mathcal{A}}_{\tau}^{-1}} \leq \lambda_{1} \left \| \boldsymbol{\Theta}_{*} \right \| \leq \lambda_{1} \beta_{\theta}
\end{displaymath}

By integrating the above three terms, we complete the proof of the bound for $\alpha_{\theta}$ in \cref{lemma:bound}. The proof for the bound of $\alpha_{u}$ follows a similar approach and is therefore omitted.

\end{proof}

\subsection{Proof of \cref{theorem:regret}}
\label{app:theorem_3_5}
In this subsection, for description conciseness, the subscripts of functions are omitted when there is no risk of confusion. For example, we denote $r_{\tau +1} (\boldsymbol{b}_{\tau + 1})$ simply as $r(\boldsymbol{b}_{\tau + 1})$. Besides, we denote the acceptance estimated by $\bar{r}_{\tau + 1} (\cdot) + e_{\tau +1} (\cdot)$ as $s(\cdot)$, and the samples as $\boldsymbol{\mu}_{\tau+1} = \phi(\boldsymbol{b}_{\tau + 1} \boldsymbol{Y}) \otimes \phi(\boldsymbol{x}_{\tau + 1})$ and $\boldsymbol{v}_{\tau+1} = \phi(\boldsymbol{b}_{\tau + 1} \boldsymbol{Y}) \otimes \boldsymbol{p}_{\tau + 1}$. Additionally, the optimal bid at time step $\tau+1$ is denoted as $\boldsymbol{b}_{\tau + 1}^{*}$, thus $r(\boldsymbol{b}_{\tau + 1}^{*})$ and $r(\boldsymbol{b}_{\tau + 1})$ are the true acceptance of the optimal bid $\boldsymbol{b}_{\tau + 1}^{*}$ and the chosen bid $\boldsymbol{b}_{\tau + 1}$ at $\tau + 1$, respectively.

\begin{proof}
    Firstly, we analyze \autoref{eq:ucb_kernel}.
    \begin{displaymath}
        \begin{split}
            & \boldsymbol{\mathcal{A}}_{\tau} (\phi(\boldsymbol{b}_{\tau + 1} \boldsymbol{Y}) \otimes \phi(\boldsymbol{x}_{\tau + 1}))^{\mathsf{T}} = (\boldsymbol{A}_{\tau}^{\mathsf{T}} \boldsymbol{A}_{\tau} + \lambda_{1} \boldsymbol{I}_{h^{2}}) \boldsymbol{\mu}_{\tau +1}^{\mathsf{T}} = \boldsymbol{A}_{\tau}^{\mathsf{T}}  \bar{\boldsymbol{k}}_{\tau + 1}^{\mathsf{T}} + \lambda_{1} \boldsymbol{\mu}_{\tau + 1}^{\mathsf{T}}
        \end{split}
    \end{displaymath}

    Rearranging the above equation, there is:
    \begin{displaymath}
        \begin{split}
            \boldsymbol{\mu}_{\tau +1}^{\mathsf{T}} = & \boldsymbol{\mathcal{A}}_{\tau}^{-1} (\boldsymbol{A}_{\tau}^{\mathsf{T}} \bar{\boldsymbol{k}}_{\tau + 1}^{\mathsf{T}} + \lambda_{1} \boldsymbol{\mu}_{\tau + 1}^{\mathsf{T}}) \\ 
            = & \boldsymbol{A}_{\tau}^{\mathsf{T}} (\boldsymbol{K}_{\tau} + \lambda_{1} \boldsymbol{I}_{\tau})^{-1} \bar{\boldsymbol{k}}_{\tau + 1}^{\mathsf{T}} + \lambda_{1} \boldsymbol{\mathcal{A}}_{\tau}^{-1} \boldsymbol{\mu}_{\tau +1}^{\mathsf{T}}
        \end{split}
    \end{displaymath}

    The last equality above is based on the study of Haasdonk et al.~\cite{kernel}. As the kernel value at time step $\tau +1$ is denoted as $k_{\tau +1} = \boldsymbol{\mu}_{\tau +1} \boldsymbol{\mu}_{\tau +1}^{\mathsf{T}}$, there is:
    
    \begin{displaymath}
        \begin{split}
            & \boldsymbol{\mu}_{\tau +1} \boldsymbol{\mu}_{\tau +1}^{\mathsf{T}} = \bar{\boldsymbol{k}}_{\tau + 1} (\boldsymbol{K}_{\tau} + \lambda_{1} \boldsymbol{I}_{\tau})^{-1} \bar{\boldsymbol{k}}_{\tau + 1}^{\mathsf{T}} + \lambda_{1} \textcolor{red}{\boldsymbol{\mu}_{\tau +1} \boldsymbol{\mathcal{A}}_{\tau}^{-1} \boldsymbol{\mu}_{\tau +1}^{\mathsf{T}}} \\ 
            \Rightarrow & \textcolor{red}{\boldsymbol{\mu}_{\tau +1} \boldsymbol{\mathcal{A}}_{\tau}^{-1} \boldsymbol{\mu}_{\tau +1}^{\mathsf{T}}} = \frac{1}{\lambda_1} ( k_{\tau + 1} - \bar{\boldsymbol{k}}_{\tau + 1} (\boldsymbol{K}_{\tau} + \lambda_{1} \boldsymbol{I}_{\tau})^{-1} \bar{\boldsymbol{k}}_{\tau + 1}^{\mathsf{T}})
        \end{split}
    \end{displaymath}
    
    Derivation for $\boldsymbol{v}_{\tau +1} \boldsymbol{\mathcal{D}}_{\tau}^{-1} \boldsymbol{v}_{\tau +1}^{\mathsf{T}}$ is similar. From the above results, \autoref{eq:ucb_kernel} is equivalent to the following format, consistent with existing UCB-based approaches. The remaining proof is based on this result.

    \begin{displaymath}
        \begin{split}
            e_{\tau + 1} = & \alpha_{\theta} \sqrt{\textcolor{red}{\boldsymbol{\mu}_{\tau +1} \boldsymbol{\mathcal{A}}_{\tau}^{-1} \boldsymbol{\mu}_{\tau +1}^{\mathsf{T}}}} + \alpha_{u} \sqrt{\textcolor{blue}{\boldsymbol{v}_{\tau +1} \boldsymbol{\mathcal{D}}_{\tau}^{-1} \boldsymbol{v}_{\tau +1}^{\mathsf{T}}}}
        \end{split}
    \end{displaymath}
    
    Secondly, we prove that $s(\boldsymbol{b}_{\tau + 1}^{*}) \geq r(\boldsymbol{b}_{\tau + 1}^{*})$.

    \begin{displaymath}
    \begin{split}
        s(\boldsymbol{b}_{\tau + 1}^{*}) - r(\boldsymbol{b}_{\tau + 1}^{*}) = & \boldsymbol{\mu}_{\tau +1}^{*} (\text{vec}(\boldsymbol{\Theta}_{\tau}) - \text{vec}(\boldsymbol{\Theta}_{*})) + \boldsymbol{v}_{\tau +1}^{*} (\text{vec}(\boldsymbol{U}_{\tau}) - \text{vec}(\boldsymbol{U}_{*})) + \alpha_{\theta} \left \| \boldsymbol{\mu}_{\tau +1}^{*} \right \|_{\boldsymbol{\mathcal{A}}_{\tau}^{-1}} + \alpha_{u} \left \| \boldsymbol{v}_{\tau +1}^{*} \right \|_{\boldsymbol{\mathcal{D}}_{\tau}^{-1}} \\ 
        \geq & - \left \| \text{vec}(\boldsymbol{\Theta}_{\tau}) - \text{vec}(\boldsymbol{\Theta}_{*}) \right \|_{\boldsymbol{\mathcal{A}}_{\tau}} \left \| \boldsymbol{\mu}_{\tau +1}^{*} \right \|_{\boldsymbol{\mathcal{A}}_{\tau}^{-1}} - \left \| \text{vec}(\boldsymbol{U}_{\tau}) - \text{vec}(\boldsymbol{U}_{*}) \right \|_{\boldsymbol{\mathcal{D}}_{\tau}} \left \| \boldsymbol{v}_{\tau +1}^{*} \right \|_{\boldsymbol{\mathcal{D}}_{\tau}^{-1}} \\ 
        & + \alpha_{\theta} \left \| \boldsymbol{\mu}_{\tau +1}^{*} \right \|_{\boldsymbol{\mathcal{A}}_{\tau}^{-1}} + \alpha_{u} \left \| \boldsymbol{v}_{\tau +1}^{*} \right \|_{\boldsymbol{\mathcal{D}}_{\tau}^{-1}} \\
        \geq & - \alpha_{\theta} \left \| \boldsymbol{\mu}_{\tau +1}^{*} \right \|_{\boldsymbol{\mathcal{A}}_{\tau}^{-1}} - \alpha_{u} \left \| \boldsymbol{v}_{\tau +1}^{*} \right \|_{\boldsymbol{\mathcal{D}}_{\tau}^{-1}} + \alpha_{\theta} \left \| \boldsymbol{\mu}_{\tau +1}^{*} \right \|_{\boldsymbol{\mathcal{A}}_{\tau}^{-1}} + \alpha_{u} \left \| \boldsymbol{v}_{\tau +1}^{*} \right \|_{\boldsymbol{\mathcal{D}}_{\tau}^{-1}} = 0
    \end{split}
\end{displaymath}

Thirdly, we bound $r(\boldsymbol{b}_{\tau + 1}^{*}) \times f(\boldsymbol{b}_{\tau + 1}^{*}) - r(\boldsymbol{b}_{\tau + 1}) \times f(\boldsymbol{b}_{\tau + 1})$. As $\boldsymbol{b}_{\tau +1}$ is the bid chosen by the NegUCB algorithm at time step $\tau + 1$, we have:

    \begin{displaymath}
    \begin{split}
        & r(\boldsymbol{b}_{\tau + 1}^{*}) \times f(\boldsymbol{b}_{\tau + 1}^{*}) \leq s(\boldsymbol{b}_{\tau + 1}^{*})  \times f(\boldsymbol{b}_{\tau + 1}^{*}) \leq s(\boldsymbol{b}_{\tau + 1})  \times f(\boldsymbol{b}_{\tau + 1}) \\[10pt]
        \Rightarrow & r(\boldsymbol{b}_{\tau + 1}^{*}) \times f(\boldsymbol{b}_{\tau + 1}^{*}) - r(\boldsymbol{b}_{\tau + 1}) \times f(\boldsymbol{b}_{\tau + 1}) \\[10pt]
        \leq & \left \{ \boldsymbol{\mu}_{\tau +1} \text{vec}(\boldsymbol{\Theta}_{\tau}) + \boldsymbol{v}_{\tau +1} \text{vec}(\boldsymbol{U}_{\tau}) + \alpha_{\theta} \left \| \boldsymbol{\mu}_{\tau +1} \right \|_{\boldsymbol{\mathcal{A}}_{\tau}^{-1}} + \alpha_{u} \left \| \boldsymbol{v}_{\tau +1} \right \|_{\boldsymbol{\mathcal{D}}_{\tau}^{-1}} - \boldsymbol{\mu}_{\tau +1} \text{vec}(\boldsymbol{\Theta}_{*}) - \boldsymbol{v}_{\tau +1} \text{vec}(\boldsymbol{U}_{*}) \right \} \times f(\boldsymbol{b}_{\tau + 1}) \\
        = & \left \{ \boldsymbol{\mu}_{\tau +1} (\text{vec}(\boldsymbol{\Theta}_{\tau}) - \text{vec}(\boldsymbol{\Theta}_{*})) + \boldsymbol{v}_{\tau +1} (\text{vec}(\boldsymbol{U}_{\tau}) - \text{vec}(\boldsymbol{U}_{*})) + \alpha_{\theta} \left \| \boldsymbol{\mu}_{\tau +1} \right \|_{\boldsymbol{\mathcal{A}}_{\tau}^{-1}} + \alpha_{u} \left \| \boldsymbol{v}_{\tau +1} \right \|_{\boldsymbol{\mathcal{D}}_{\tau}^{-1}} \right \} \times f(\boldsymbol{b}_{\tau + 1}) \\
        \leq & \left \{ 2 \alpha_{\theta} \left \| \boldsymbol{\mu}_{\tau +1} \right \|_{\boldsymbol{\mathcal{A}}_{\tau}^{-1}} + 2 \alpha_{u} \left \| \boldsymbol{v}_{\tau +1} \right \|_{\boldsymbol{\mathcal{D}}_{\tau}^{-1}} \right \} \times f(\boldsymbol{b}_{\tau + 1})
    \end{split}
    \end{displaymath}

The first inequality above is from the conclusion of the second proof step. Lastly, we prove the bound of cumulative regret. For the benefit function $f_{t}$ at time step $t$, we assume an union bound $\alpha_{f}$ such that $|f_{t}| \leq \alpha_{f}$ for $\forall \boldsymbol{b} \in B_{t}$ and $\forall t \in \left \{ 1, 2, ..., \tau \right \}$.

\begin{displaymath}
    \begin{split}
        & \sum_{t=0}^{\tau} r(\boldsymbol{b}_{t +1}^{*}) \times f(\boldsymbol{b}_{\tau + 1}^{*}) - r(\boldsymbol{b}_{t +1}) \times f(\boldsymbol{b}_{\tau + 1}) \\
        \leq & 2 \alpha_{\theta} \alpha_{f} \sum_{t=0}^{\tau} \left \| \phi(\boldsymbol{b}_{t+1} \boldsymbol{Y}) \otimes \phi(\boldsymbol{x}_{t+1}) \right \|_{\boldsymbol{\mathcal{A}}_{\tau}^{-1}} + 2 \alpha_{u} \alpha_{f} \sum_{t=0}^{\tau} \left \| \phi(\boldsymbol{b}_{t+1} \boldsymbol{Y}) \otimes \boldsymbol{p}_{t+1} \right \|_{\boldsymbol{\mathcal{D}}_{\tau}^{-1}} \\
        \leq & 2 \alpha_{\theta} \alpha_{f} \sqrt{\tau \sum_{t=0}^{\tau} \left \| \phi(\boldsymbol{b}_{t+1} \boldsymbol{Y}) \otimes \phi(\boldsymbol{x}_{t+1}) \right \|_{\boldsymbol{\mathcal{A}}_{t}^{-1}}^{2}} + 2 \alpha_{u} \alpha_{f} \sqrt{\tau \sum_{t=0}^{\tau} \left \| \phi(\boldsymbol{b}_{t+1} \boldsymbol{Y}) \otimes \boldsymbol{p}_{t+1} \right \|_{\boldsymbol{\mathcal{D}}_{t}^{-1}}^{2}} \\
        \leq & 2 \alpha_{\theta} \alpha_{f} \sqrt{2 \tau \text{log} \frac{\text{det} (\boldsymbol{\mathcal{A}}_{\tau})}{\text{det} (\lambda_{1} \boldsymbol{I}_{h^{2}})}} + 2 \alpha_{u} \alpha_{f} \sqrt{2 \tau \text{log} \frac{\text{det} (\boldsymbol{\mathcal{D}}_{\tau})}{\text{det} (\lambda_{2} \boldsymbol{I}_{mh})}} \\
        \leq & 2 \alpha_{\theta} \alpha_{f} \sqrt{2 \tau h_{*} \text{log} (1 + \frac{\tau}{\lambda_{1} h_{*}})} + 2 \alpha_{u} \alpha_{f} \sqrt{2 \tau m_{*} \text{log} (1 + \frac{\tau}{\lambda_{2} m_{*}})} \\
    \end{split}
\end{displaymath}

The third inequality is based on Lemma 11 of Abbasi-Yadkori's study~\cite{linear_bandit}, and the last inequality is based on Determinant-trace inequality to both $\bar{\boldsymbol{\mathcal{A}}_{\tau}}$ and $\bar{\boldsymbol{\mathcal{D}}}_{\tau}$. We do not explicitly emphasize the valid bid set $B_{t}$ in the above proof. However, the proof process remains the same if incorporating $B_{t}$.

\end{proof}

\subsubsection{Cumulative Regret Analysis}
\label{app:cumulative_regret_analysis}
The cumulative regret upper bound of NegUCB remains \textbf{independent of the cardinality of bids}, a notable distinction from existing algorithms like C2UCB~\cite{C2UCB}, ComLinUCB~\cite{ComLinUCB}, CC-MAB~\cite{CC-MAB}, etc., which exhibit upper bounds that are sub-linear concerning the cardinality of super arms. This behavior is attributed to the full-bandit feedback of negotiation problems and the \cref{ass:psi}. For instance, ComLinUCB estimates the reward of each arm, from which the general rewards of super arms are calculated, leading to error propagation. In contrast, NegUCB directly estimates the general rewards of super arms, eliminating error propagation. \cref{ass:psi} requires that item contexts are defined by their basic features. Neglecting this principle may result in inaccuracies when capturing bid contexts. However, similar assumptions are commonly used in various applications, such as \textit{recommendation} and \textit{crowdsourcing}. Addressing this limitation in future research is encouraged. For full-bandit feedback, existing works such as DART~\cite{DART}, ETCG~\cite{ETCG}, RGL~\cite{RGL}, etc., have been developed. However, their regret upper bounds are contingent on the cardinality of super arms, as they lack consideration of contexts.

\section{Experiment}
\label{app:experiment}

\subsection{Baseline Selection}
\label{app:baseline_seletion}
Algorithms designed for contextual combinatorial bandits include C2UCB~\cite{C2UCB}, ComLinUCB~\cite{ComLinUCB}, CC-MAB~\cite{CC-MAB}, CN-UCB~\cite{CN-UCB}, and others. However, these algorithms are tailored for semi-bandit feedback and cannot be directly applied to our specific problems. Our experiments adapt LinUCB~\cite{LinUCB} to combinatorial bandits with full-bandit feedback. Specifically, we extend LinUCB by incorporating \cref{ass:psi}, which treats bids as the basic arms of the original algorithm. The algorithms DART~\cite{DART}, ETCG~\cite{ETCG}, and RGL~\cite{RGL} are specifically tailored for full-bandit feedback. However, they do not consider contextual information, rendering them unsuitable for addressing our particular problems. 

Neural network-based algorithms, e.g., Neural-UCB, CN-UCB, or variants, are not chosen as the baselines in our experiments. Although neural networks have powerful representation capabilities, the networks in neural bandits cannot be large, as the computational complexity is cubic concerning the number of network parameters, limiting their capabilities. Neural-LinUCB~\cite{shallow_exploration} solely explores the output layer to expedite the neural-bandit algorithms. Nevertheless, it encounters instability issues in the iteration process among three components: learning the parameters $\boldsymbol{\Theta}$, $\boldsymbol{U}$, and learning the neural network $\phi$. According to exiting works~\cite{TCB}, prior knowledge or additional constraints are required to govern learning under these scenarios.

In addition to bandit-based algorithms, alternative methods for negotiation have been proposed~\cite{Kris_Cao, MCTS, Anegma}. However, as discussed earlier, these approaches often struggle to address the exploitation-exploration dilemma and handle large action spaces effectively. Consequently, they are anticipated to exhibit inferior performance compared to NegUCB. In our experiments, we adopt a variant of ANEGMA \cite{Anegma}. The notable difference is the absence of the pre-training component, caused by the unavailability of historical negotiation data for various negotiation tasks. To compensate for this omission, we extend the training duration of ANEGMA, ensuring that the results accurately reflect its true capabilities.

\subsection{Bid Design}
\label{app:bid_design}

In this subsection, we illustrate the design of bid vectors for each task through illustrative examples. These bids are structured based on the NegUCB algorithm and tailored to specific negotiation problems. However, it's advisable to flexibly adjust bid formats to accommodate diverse problems and algorithms.

\subsubsection{Multi-issue Negotiation}
Consider a negotiation scenario with four issues, each having $4, 2, 2, 3$ possible values, respectively. The bid vector, representing the potential outcomes for these issues, is of size $4 + 2 + 2 + 3 = 11$. For instance, a bid expressing \textit{value 3} for issue A, \textit{value 1} for issue B, \textit{value 2} for issue C, and \textit{value 2} for issue D is denoted by the bid vector $\boldsymbol{b} = (0, 0, 1, 0, 1, 0, 0, 1, 0, 1, 0)$. It signifies the selection of values for each issue.

\subsubsection{Resource Allocation}
\cref{fig:NegUCB} provides a simple example of how a bid can be designed in a resource allocation task, where each negotiator takes distinct categories of items. However, both negotiators may want some items in the same category. For example, both of them want some apples. Consequently, the bids for this task in our experiment are designed in a more general way. Let's assume three types of items are available: 4 strawberries, 2 peppers, and 5 apples. A bid representing a request for 1 strawberry, 1 pepper, and 2 apples, while the counterpart negotiator retains the remaining 3 strawberries, 1 pepper, and 3 apples, can be denoted by the bid vector $\boldsymbol{b} = (1, 1, 2, -3, -1, -3)$. The first three entries represent the items requested by our negotiator, while the remaining three represent those for the counterpart. Under this setting, the item context matrix is:

\vspace{-10pt}
\begin{tiny}
\begin{align}
\boldsymbol{Y} = \begin{pmatrix}
2 & 1 \\ 
1 & 3 \\ 
5 & 4 \\ 
--- & --- \\
2 & 1 \\ 
1 & 3 \\ 
5 & 4
\end{pmatrix}
\label{eq:Y}
\end{align}
\end{tiny}

\subsubsection{Trading}
For a trading scenario involving three types of items—strawberries, peppers, and apples—a bid indicating that our negotiator offers 1 pepper and 1 strawberry to the counterpart while seeking 2 apples in return can be represented by the bid vector $\boldsymbol{b} = (1, 1, 0, 0, 0, -2)$. The first three entries signify the items provided by our negotiator, whereas the remaining three entries denote the items sought from the counterpart. The item context matrix is similar to that in \autoref{eq:Y}.

\subsection{More Experiment Results and Analysis to \textit{Resource Allocation}}
\label{app:resource_allocation}

\begin{table}
\caption{Exploration parameter settings for each algorithm.}
\label{tab:exploration_rate}
\vspace{5pt}
\begin{center}
\scalebox{0.8}{

    \begin{tabular}{cccccccc}
        \toprule
        \multicolumn{2}{c}{Algorithm}  & $\alpha_{1}$ & $\alpha_{2}$ & $\alpha_{3}$ & $\alpha_{4}$ & $\alpha_{5}$ & $\alpha_{6}$ \\
        \midrule
        \multicolumn{2}{c}{LinUCB} & 0 & 1 & 4 & \textbf{8} & 16 & 32 \\
        \multicolumn{2}{c}{ANAC agent}  & 0 & \textbf{1} & 2 & 3 & 4 & 5 \\
        \hline
        \multirow{2}*{FactorUCB} & \# items is $5$  & 0 & 0.4 & \textbf{0.8} & 1.2 & 1.6 & 2 \\
        & \# items is $20$ & 0 & \textbf{0.1} & 0.4 & 0.6 & 0.8 & 1 \\
        \hline
        \multicolumn{2}{c}{KernelUCB}  & 0 & \textbf{1} & 2 & 4 & 6 & 8 \\
        \multicolumn{2}{c}{NegUCB}  & 0 & \textbf{0.1} & 0.4 & 0.6 & 0.8 & 1 \\    
       \bottomrule
   \end{tabular}

}

\end{center}
 \end{table}

\begin{figure*}[t!]
     \centering
     \subfigure[Acceptance by reinforcement learning]{
     \includegraphics[width=0.31\textwidth, trim={0 0 1.5cm 1.5cm}, clip]{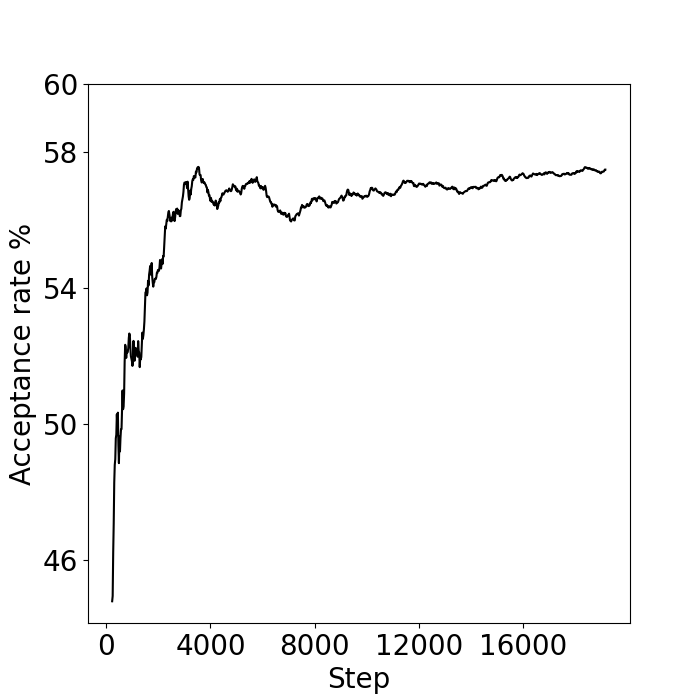}
     \label{fig:rl_acceptance_appendix}
     }
     \hfill
     \subfigure[$\times$ 100: Cumulative acceptance regret]{
     \includegraphics[width=0.31\textwidth, trim={0 0 1.5cm 1.5cm}, clip]{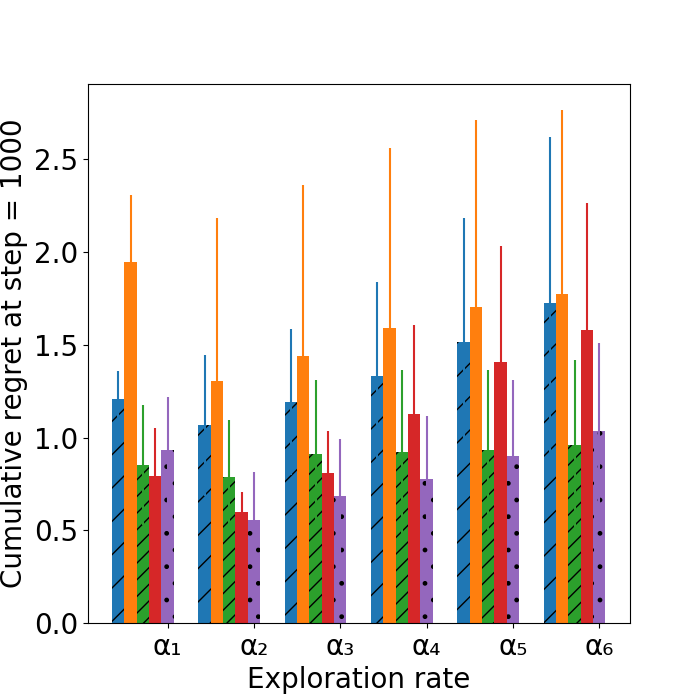}
     \label{fig:syn_exploration_rate_appendix}
     }
     \subfigure[$\times 100$: Cumulative acceptance regret]{
     \includegraphics[width=0.31\textwidth, trim={0 0 1.5cm 1.5cm}, clip]{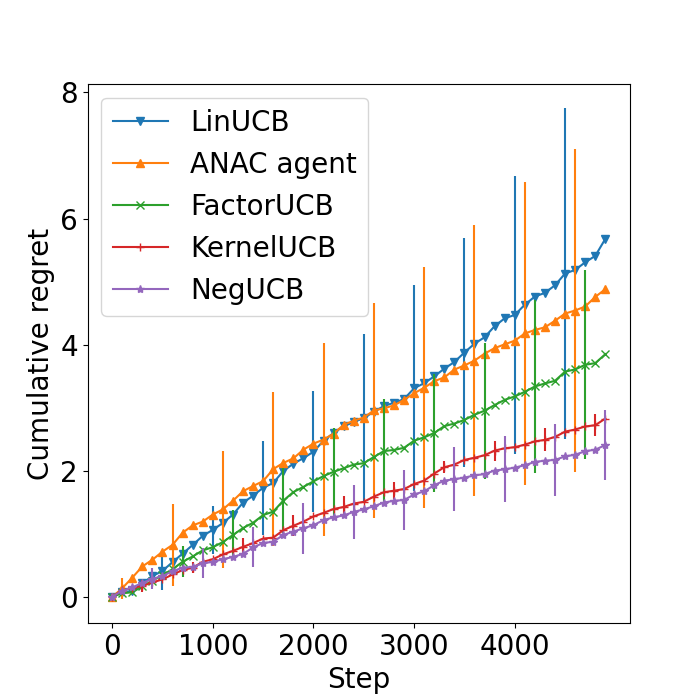}
     \label{fig:syn_acceptance_rate_appendix}
     }
     \caption{More experiment results on resource allocation task.}
\end{figure*}

Because the exploitation scales of algorithms vary, their exploration scales also vary largely. We conduct a search and summarize the six representative exploration rates corresponding to \cref{fig:syn_1} in \cref{tab:exploration_rate}, where the rates in bold are the optimal ones of each algorithm. \cref{fig:rl_acceptance_appendix} presents the outcomes of the reinforcement learning method. Due to the random nature of exploration in reinforcement learning, employing $\epsilon$-greedy exploration, we extend the training duration to $20000$ steps to ensure the results accurately depict the algorithm's true capabilities. As depicted in \cref{fig:rl_acceptance_appendix}, the acceptance rate exhibits an initial increase but soon plateaus, facing challenges in surpassing an acceptance rate of $0.6$.

In \cref{fig:syn_acceptance_rate}, the acceptance rates exhibit initial fluctuations due to the limited negotiation steps, resulting in sharp increases when deals occur, primarily caused by randomness. It is important to note that these initial spikes do not necessarily imply high negotiation capabilities. Other studies have documented similar observations, such as TCB~\cite{TCB} and FactorUCB~\cite{FactorUCB}.

Additionally, we add one more experiment with a larger action space. Assuming there are three categories of items and the number of items in each category does not exceed $20$, the action space comprises at most $21 \times 21 \times 21 = 9261$ actions. Experiment results under this setting are shown in \cref{fig:syn_exploration_rate_appendix}, \cref{fig:syn_acceptance_rate_appendix}, and \cref{fig:rl_acceptance_appendix_}, demonstrating the advantages of NegUCB compared to the baselines. It is worth noting that the complexity of the task is not solely determined by the size of the action space but also by other variables, such as the item contexts and the attitudes of the counterparts, which may be the reason why the acceptance rates of the added experiment are higher than those in the main content. 

\subsection{More Experiment Results and Analysis to \textit{Trading}}
\label{app:trading}

\begin{figure*}[t!]
     \centering
     \subfigure[Acceptance by reinforcement learning]{
     \includegraphics[width=0.31\textwidth, trim={0 0 1.5cm 1.5cm}, clip]{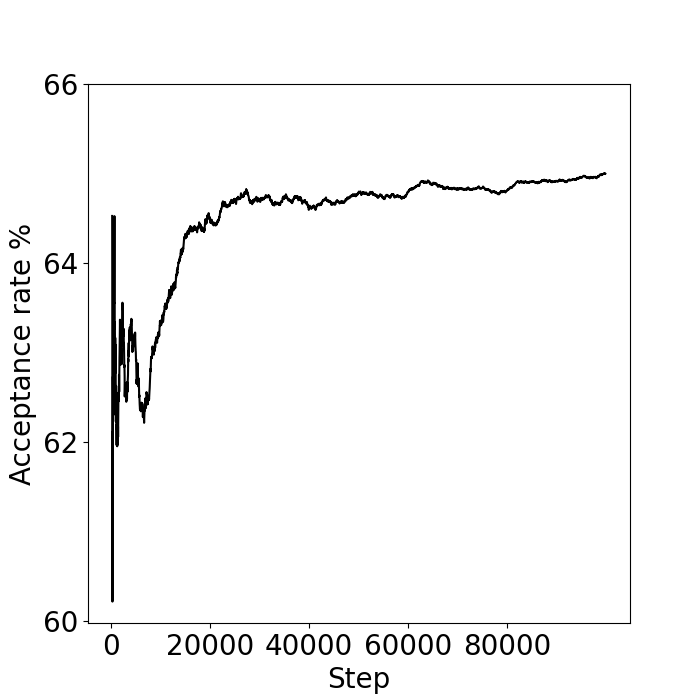}
     \label{fig:rl_acceptance_appendix_}
     }
     \hfill
     \subfigure[Deal rate w.r.t. exploration rate]{
     \includegraphics[width=0.31\textwidth, trim={0 0 1.5cm 1.5cm}, clip]{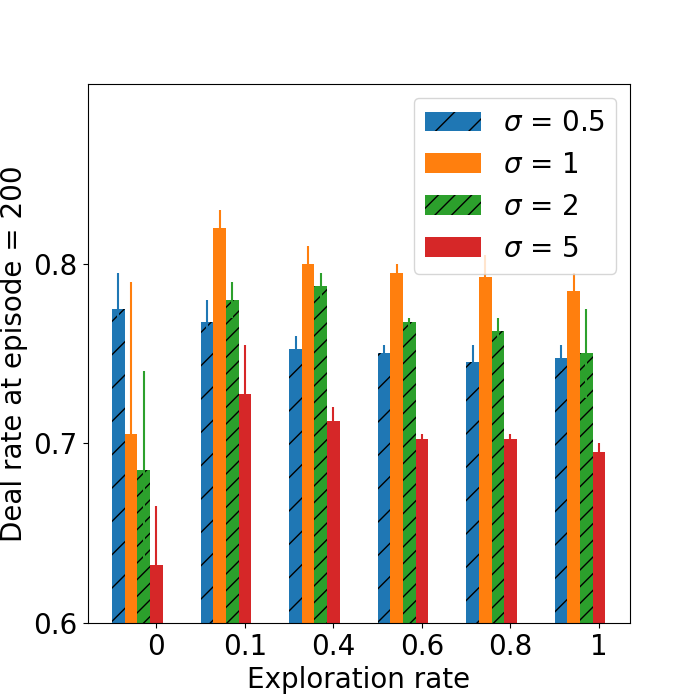}
     \label{fig:civ_exploration_rate}
     }
     \subfigure[Acceptance rate]{
     \includegraphics[width=0.31\textwidth, trim={0 0 1.5cm 1.5cm}, clip]{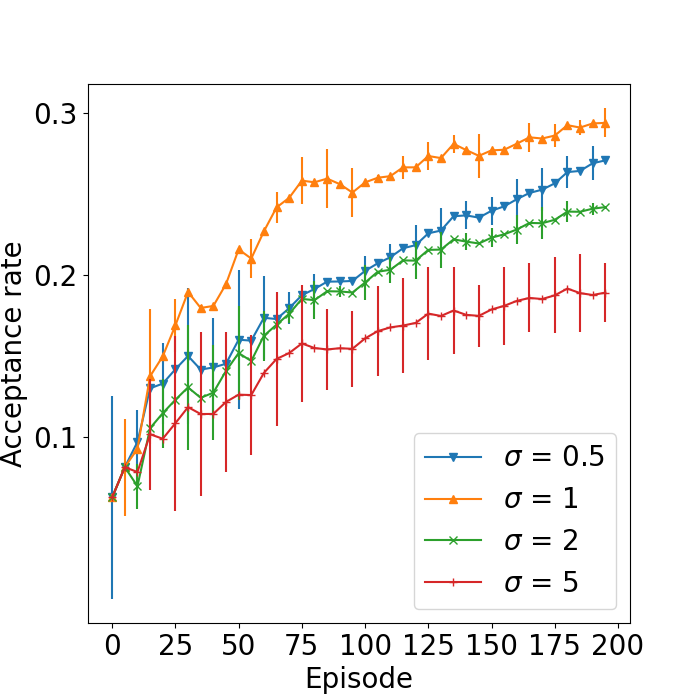}
     \label{fig:civ_acceptance_app}
     }
     \caption{More experiment results on resource allocation and trading tasks. \textit{Deal rate} is a metric defined by CivRealm, quantifying the percentage of episodes that result in a deal, while the \textit{acceptance rate} is the percentage of the proposed bids being accepted..}
\end{figure*}

In the CivRealm experiment, we have chosen a bid cardinality of $\gamma = 4$ for the sake of experiment efficiency. Setting $\gamma$ too large would require additional search algorithms to find the optimal bid in \autoref{eq:bid eq.}, which falls beyond the scope of this work and introduces errors unrelated to our algorithm. Moreover, bids in \textit{trading} often consist of only a few items. However, this is not contradictory to the large action space issue, as the action space has a cardinality of $\sum_{j=1}^{\gamma} \binom{n}{j}$, which can still be large even for a small $\gamma$.

The contexts of negotiator pairs encompass the technologies our negotiator and the counterpart possess. Contexts of technologies include metrics such as \textit{cost}, \textit{research\_reqs\_count}, and \textit{num\_reqs}, which are provided by CivRealm \cite{CivRealm} and describe fundamental features of technologies.

In this experiment, we employ the SE kernel given by $\kappa(\boldsymbol{x}_{w}, \boldsymbol{x}_{j}) = \text{exp} (- \frac{1}{2 \sigma ^{2}} \left \| \boldsymbol{x}_{w} - \boldsymbol{x}_{j} \right \|^{2})$. We systematically explore SE kernels with diverse hyper-parameters $\sigma$, specifically $\sigma = 0.5, 1, 2$, and $5$, to fine-tune the most suitable kernel function for the trading task discussed in this subsection. The results, illustrated in \cref{fig:civ_exploration_rate}, display the final deal rates after $200$ episodes using different kernel functions. According to the experiment results from CivRealm \cite{CivRealm}, their deal rates are less than $0.4$, notably inferior to those of NegUCB. Based on \cref{fig:civ_exploration_rate}, we conclude that the SE kernel with $\sigma=1$ emerges as the most suitable choice for the trading task on CivRealm. Additionally, \cref{fig:civ_acceptance_app} further illustrates the acceptance rates of NegUCB with various SE kernels under their corresponding optimal exploration rates, specifically $\alpha_{\theta} = \alpha_{u} = 0, 0.1, 0.4, 0.1$ for $\sigma = 0.5, 1, 2, 5$, respectively. 

\end{document}